%% file: main.tex
\documentclass[10pt,journal,compsoc]{IEEEtran}
\ifCLASSOPTIONcompsoc
  \usepackage[nocompress]{cite}
\else
  \usepackage{cite}
\fi
\ifCLASSINFOpdf
   \usepackage[pdftex]{graphicx}
\else
\fi
\usepackage{tikz}

\usepackage{amsmath, amsthm, amssymb, amsfonts}

\usepackage[ruled,vlined]{algorithm2e}

\DeclareMathOperator{\Tr}{tr}

\DeclareMathOperator*{\argmin}{arg\,min}

\usepackage{parskip}

\theoremstyle{plain}
\newtheorem{theorem}{Theorem}
\newtheorem{lemma}[theorem]{Lemma}

\usepackage{algorithmicx}
\usepackage{algpseudocode}
\usepackage[caption=false,font=footnotesize,labelfont=sf,textfont=sf]{subfig}
\begin{document}
%
% paper title
% Titles are generally capitalized except for words such as a, an, and, as,
% at, but, by, for, in, nor, of, on, or, the, to and up, which are usually
% not capitalized unless they are the first or last word of the title.
% Linebreaks \\ can be used within to get better formatting as desired.
% Do not put math or special symbols in the title.
\title{Rotation Averaging: A Primal-Dual Method and Closed-Forms in Cycle Graphs}
%
%
% author names and IEEE memberships
% note positions of commas and nonbreaking spaces ( ~ ) LaTeX will not break
% a structure at a ~ so this keeps an author's name from being broken across
% two lines.
% use \thanks{} to gain access to the first footnote area
% a separate \thanks must be used for each paragraph as LaTeX2e's \thanks
% was not built to handle multiple paragraphs
%
%
%\IEEEcompsocitemizethanks is a special \thanks that produces the bulleted
% lists the Computer Society journals use for "first footnote" author
% affiliations. Use \IEEEcompsocthanksitem which works much like \item
% for each affiliation group. When not in compsoc mode,
% \IEEEcompsocitemizethanks becomes like \thanks and
% \IEEEcompsocthanksitem becomes a line break with idention. This
% facilitates dual compilation, although admittedly the differences in the
% desired content of \author between the different types of papers makes a
% one-size-fits-all approach a daunting prospect. For instance, compsoc 
% journal papers have the author affiliations above the "Manuscript
% received ..."  text while in non-compsoc journals this is reversed. Sigh.

\author{Gabriel~Moreira,
        Manuel~Marques,
        and~João~Paulo~Costeira% <-this % stops a space
\IEEEcompsocitemizethanks{\IEEEcompsocthanksitem G Moreira is with the LTI, Carnegie Mellon University, Pittsburgh, PA 15213 and with the Institute for Systems and Robotics, Instituto Superior Técnico, Lisboa 1049-001, Portugal. E-mail: gmoreira@cs.cmu.edu\protect\\
% note need leading \protect in front of \\ to get a newline within \thanks as
% \\ is fragile and will error, could use \hfil\break instead.
\IEEEcompsocthanksitem M Marques and JP Costeira are with the Institute for Systems and Robotics, Instituto Superior Técnico, Lisboa 1049-001, Portugal.}% <-this % stops an unwanted space
%\thanks{Manuscript received April 19, 2005; revised August 26, 2015.}
}

% note the % following the last \IEEEmembership and also \thanks - 
% these prevent an unwanted space from occurring between the last author name
% and the end of the author line. i.e., if you had this:
% 
% \author{....lastname \thanks{...} \thanks{...} }
%                     ^------------^------------^----Do not want these spaces!
%
% a space would be appended to the last name and could cause every name on that
% line to be shifted left slightly. This is one of those "LaTeX things". For
% instance, "\textbf{A} \textbf{B}" will typeset as "A B" not "AB". To get
% "AB" then you have to do: "\textbf{A}\textbf{B}"
% \thanks is no different in this regard, so shield the last } of each \thanks
% that ends a line with a % and do not let a space in before the next \thanks.
% Spaces after \IEEEmembership other than the last one are OK (and needed) as
% you are supposed to have spaces between the names. For what it is worth,
% this is a minor point as most people would not even notice if the said evil
% space somehow managed to creep in.

% The paper headers
\markboth{Journal of \LaTeX\ Class Files,~Vol.~XX, No.~XXX, XXX}%
{Shell \MakeLowercase{\textit{et al.}}: Bare Demo of IEEEtran.cls for Computer Society Journals}
% The only time the second header will appear is for the odd numbered pages
% after the title page when using the twoside option.
% 
% *** Note that you probably will NOT want to include the author's ***
% *** name in the headers of peer review papers.                   ***
% You can use \ifCLASSOPTIONpeerreview for conditional compilation here if
% you desire.

% The publisher's ID mark at the bottom of the page is less important with
% Computer Society journal papers as those publications place the marks
% outside of the main text columns and, therefore, unlike regular IEEE
% journals, the available text space is not reduced by their presence.
% If you want to put a publisher's ID mark on the page you can do it like
% this:
%\IEEEpubid{0000--0000/00\$00.00~\copyright~2015 IEEE}
% or like this to get the Computer Society new two part style.
%\IEEEpubid{\makebox[\columnwidth]{\hfill 0000--0000/00/\$00.00~\copyright~2015 IEEE}%
%\hspace{\columnsep}\makebox[\columnwidth]{Published by the IEEE Computer Society\hfill}}
% Remember, if you use this you must call \IEEEpubidadjcol in the second
% column for its text to clear the IEEEpubid mark (Computer Society jorunal
% papers don't need this extra clearance.)

% use for special paper notices
%\IEEEspecialpapernotice{(Invited Paper)}

% for Computer Society papers, we must declare the abstract and index terms
% PRIOR to the title within the \IEEEtitleabstractindextext IEEEtran
% command as these need to go into the title area created by \maketitle.
% As a general rule, do not put math, special symbols or citations
% in the abstract or keywords.
\IEEEtitleabstractindextext{%
\begin{abstract}
A cornerstone of geometric reconstruction, rotation averaging seeks the set of absolute rotations that optimally explains a set of measured relative orientations between them. In addition to being an integral part of bundle adjustment and structure-from-motion, the problem of synchronizing rotations also finds applications in visual simultaneous localization and mapping, where it is used as an initialization for iterative solvers, and camera network calibration. Nevertheless, this optimization problem is both non-convex and high-dimensional. In this paper, we address it from a maximum likelihood estimation standpoint and make a twofold contribution. Firstly, we set forth a novel primal-dual method, motivated by the widely accepted spectral initialization. Further, we characterize stationary points of rotation averaging in cycle graphs topologies and contextualize this result within spectral graph theory. We benchmark the proposed method in multiple settings and certify our solution via duality theory, achieving a significant gain in precision and performance.
\end{abstract}

% Note that keywords are not normally used for peerreview papers.
}

% make the title area
\maketitle

% To allow for easy dual compilation without having to reenter the
% abstract/keywords data, the \IEEEtitleabstractindextext text will
% not be used in maketitle, but will appear (i.e., to be "transported")
% here as \IEEEdisplaynontitleabstractindextext when the compsoc 
% or transmag modes are not selected <OR> if conference mode is selected 
% - because all conference papers position the abstract like regular
% papers do.
\IEEEdisplaynontitleabstractindextext
% \IEEEdisplaynontitleabstractindextext has no effect when using
% compsoc or transmag under a non-conference mode.

% For peer review papers, you can put extra information on the cover
% page as needed:
% \ifCLASSOPTIONpeerreview
% \begin{center} \bfseries EDICS Category: 3-BBND \end{center}
% \fi
%
% For peerreview papers, this IEEEtran command inserts a page break and
% creates the second title. It will be ignored for other modes.
\IEEEpeerreviewmaketitle

\IEEEraisesectionheading{\section{Introduction}\label{sec:introduction}}
% Computer Society journal (but not conference!) papers do something unusual
% with the very first section heading (almost always called "Introduction").
% They place it ABOVE the main text! IEEEtran.cls does not automatically do
% this for you, but you can achieve this effect with the provided
% \IEEEraisesectionheading{} command. Note the need to keep any \label that
% is to refer to the section immediately after \section in the above as
% \IEEEraisesectionheading puts \section within a raised box.

% The very first letter is a 2 line initial drop letter followed
% by the rest of the first word in caps (small caps for compsoc).
% 
% form to use if the first word consists of a single letter:
% \IEEEPARstart{A}{demo} file is ....
% 
% form to use if you need the single drop letter followed by
% normal text (unknown if ever used by the IEEE):
% \IEEEPARstart{A}{}demo file is ....
% 
% Some journals put the first two words in caps:
% \IEEEPARstart{T}{his demo} file is ....
% 
% Here we have the typical use of a "T" for an initial drop letter
% and "HIS" in caps to complete the first word.

% - Pose Graph (...) goes to the new section
% - Comecar com a aplicação em robotics & cv bla bla e vai directa a contribuição e as coisas mais advanced.
% Criar nova seccao com as meshes de translaçao e rotação e com as explicações para amateurs, de modo a quem ja sabe poder saltar

\IEEEPARstart{P}{ose} graph optimization (PGO), synchronization over the Euclidean group and motion averaging are designations often used when referring to the problem of finding the optimal configuration of a set of vertices, given a set of corrupted relative pairwise measurements between them. What the aforementioned vertices represent varies according to the application. In Robotics, the most natural use of PGO is in Visual Simultaneous Localization and Mapping (SLAM), where each vertex corresponds to the pose of a robot at a given instant \cite{Carlone2015,Carlone2015_dual,Kummerle2011,Dellaert2012}. Variants of PGO over SO(3), known as rotation averaging \cite{hartley2013rotation}, and over the complex circle group, defined as phase synchronization \cite{Boumal2016}, have been extensively studied as well. The former usually linked to 3D reconstruction applications in Computer Vision tasks such as structure-from-motion (SfM) \cite{schonberger2016structure}, bundle-adjustment and camera network calibration \cite{Tron2008}, while the latter  finds applications in signal processing and distributed networks e.g., clock synchronization.

In this paper, we build on the MLE formulation of PGO \cite{Carlone2015_dual, Eriksson2018}, making a two-fold contribution. We characterize the stationary points in $\mathrm{SO}(p)$, for cycle graph topologies and leverage recent results on the SDP dual problem and strong duality to propose a primal-dual method with spectral updates, akin to a recursive spectral initialization \cite{Arrigoni2016}. We prove that our method is globally optimal in cycle graphs and rely on the dual SDP certificate for an empirical optimality assessment in the general case. This is accomplished via numerous experiments using well known PGO benchmark datasets \cite{Carlone2015_dual,Carlone2015}.

% sumario de contribuições
% extension of iccv 

\subsection*{An overview of PGO}
\begin{figure}
    \centering
    \includegraphics[width=\linewidth,trim={2cm 4cm 1.5cm 3.5cm},clip]{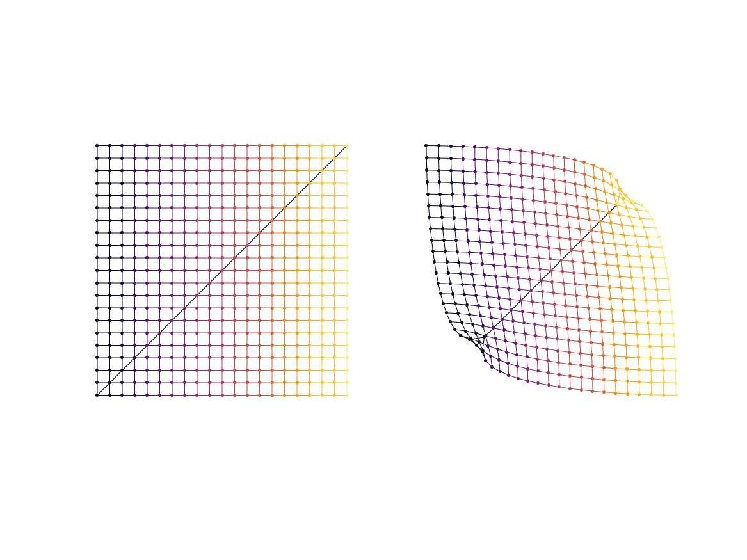}
    \caption{Left: ground-truth graph consisting of a mesh with 441 vertices. Edges of this colored grid correspond to noiseless pairwise measurements. One diagonal measurement (black) is added, which underestimates the true diagonal by 70\%. Right: optimized vertex positions.}
    % Esquerda - ground-truth measurements
    % Direita  - após optimização com diag subestimada
    \label{fig:translations_pgo}
\end{figure}

In general, the problem may be defined as follows. Let $G=(\mathcal{V},\mathcal{E})$ be a graph. Define $\mathcal{S}$ as the feasible set, which we can assume to be a group. Each pairwise tuple $(i,j) \in \mathcal{V}^2$, is identified with the composition of $x_i$ with $x_j^{-1}$ i.e., $x_{ij} := x_{i} x_j^{-1}$, which may be interpreted as the value of $i$ relative to that of $j$. Given a set of noisy measurements of these relative values $\tilde{x}_{ij},\; (i,j)\in \mathcal{E}$, PGO may be formally defined as
\begin{equation}
    \begin{aligned}
        &\underset{x_1,\dots,x_n}{\text{minimize}} & & \sum_{(i,j)\in \mathcal{E}} d
        \big(x_{i} x_j^{-1}, \tilde{x}_{ij}\big) \\
        &\text{subject to} & & x_i \in \mathcal{S},\quad i=1,\dots,n
    \end{aligned}
\label{p:general_pgo}
\end{equation}
where $d(\cdot,\cdot)$ is an appropriate distance function in $\mathcal{S}$. Consider the example of letting $\mathcal{S}$ be the group of translations $\mathbb{T}$ of $\mathbb{R}^p$. Leveraging the isomorphism between $\mathbb{T}$ and $\mathbb{R}^p$ itself and letting $d$ define the norm induced by the usual inner product of $\mathbb{R}^p$, Problem (\ref{p:general_pgo}) involves just the minimization of the least-squares cost $\sum_{(i,j)\in \mathcal{E}}\| (t_i - t_j) - \tilde{t}_{ij} \|_2^2$, whose closed-form solution is, in general, given by
\begin{equation}
    T = J^\dagger \widetilde{T},
    \label{eq:solution_euclidean_pgo}
\end{equation}
where $J$ is the signed incidence matrix of $G$, the $i$-th row of $T$ contains $t_i^\top$ and the $k$-th row of $\widetilde{T}$ contains $\tilde{t}_k^\top$. We may check that the solution given by (\ref{eq:solution_euclidean_pgo}) verifies 
\begin{equation}
    t_i^\ast = \argmin_{t\in\mathbb{R}^p} \sum_{j:(i,j)\in \mathcal{E}} \| (t-t_j^\ast) - \tilde{t}_{ij}\|_2^2,
\end{equation}
for all $i \in \mathcal{V}$. Each vertex is thus in consensus with its neighbors regarding its own position. This can be observed in Fig. \ref{fig:translations_pgo}, where we add an incorrect measurement (diagonal in black) to a noiseless graph with a mesh-like topology (on the left). The inconsistency introduced by the diagonal measurement diffuses through the graph (shown on the right), spreading between adjacent vertices. If we allow the edges of the mesh to represent springs with the equal stiffness, the solution is the state of equilibrium of the system, after the diagonal spring, with nominal length smaller than the mesh diagonal is added to the system. In this case, PGO has thus an interesting interpretation of error redistribution, which can be described from a graph consensus standpoint or in terms of an elastic spring system. 

\begin{figure}
    \centering
    \includegraphics[width=\linewidth,trim={2cm 4cm 1.5cm 3.5cm},clip]{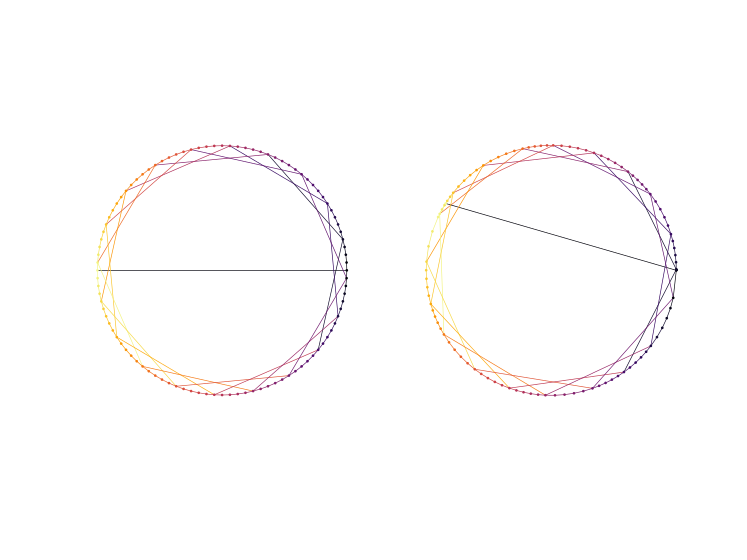}
    \caption{Left: ground-truth graph, with 100 vertices represented on the complex circle group. Colored edges correspond to noiseless measurements. The horizontal black measurement is added, with a phase error of $-\pi/4$ over the ground-truth $\pi$. Right: vertex positions of the phase synchronization global optimum.}
    \label{fig:circle_pgo}
\end{figure}

PGO's main difficulty arises when rotations are introduced, making $\mathcal{S}$ non-convex and ultimately rendering the problem less tractable. In Fig. \ref{fig:circle_pgo}, a graph over the complex circle group is shown. Once again, the colored edges correspond to noiseless measurements and the horizontal black line to an edge whose measurement has an angular error of $-\pi/4$. The problem has multiple local minima i.e., there are multiple equilibrium states for this system. The optimum is shown on the right. Remarkably, the reasoning behind our Euclidean example of Fig. \ref{fig:translations_pgo} is still valid i.e., the optimal rotations will be in consensus with their neighbors,
\begin{equation}
    R_i^\ast = \argmin_{R \in \mathrm{SO}(p)} \sum_{j:(i,j)\in \mathcal{E}} \big\| \widetilde{R}_{ij}R_j^\ast  - R\big\|_F^2,
    \label{eq:rotations_consensuus_intro}
\end{equation}
for all $i\in \mathcal{V}$. While the optimization problem in (\ref{eq:rotations_consensuus_intro}) can be solved in closed-form via a Singular Value Decomposition (SVD), the overall synchronization problem is not as tractable. In fact, under the maximum likelihood formulation pervasive in the literature, which is an instance of ($\ref{p:general_pgo}$), PGO in $\mathrm{SE}(p)$ and $\mathrm{SO}(p)$ is a variant of the max-cut problem, known as orthogonal cut and similarly to the former, it is itself NP-hard \cite{Boumal2015_riemann}. 
% - Citações
Akin to max-cut, PGO also admits a convex relaxation under the form of a Semidefinite Program (SDP), which is its bi-dual. Characterizing the tightness of this relaxation, which is desireable from an optimality certification standpoint, has been the subject of recent research \cite{Bandeira2017,Rosen2019}.

In the subsequent sections, we start by reviewing some of the recent works in PGO and general synchronization. In Section \ref{sec:problem statement}, we state our hypothesis and define the problem. We put forward our primal-dual method for the rotation synchronization problem in Section \ref{sec:primal dual} and analyze stationary points in cycle graphs in Section \ref{sec:cycle_graphs}. Empirical validation and benchmarks are present subsequently. This work is an extension of the conference paper \cite{Moreira2021}.

\section{Notation}
We use $\mathbb{R}$ and $\mathbb{C}$ to denote the real and complex numbers, respectively. The trace operator reads as $\Tr(\cdot)$. The usual inner product on $\mathbb{R}^p$ is denoted as $\langle \cdot, \cdot \rangle$ i.e., $\langle A , B \rangle = \Tr(A^\top B)$ and the operator norm thereby induced via $\| \cdot \|_2$. The Frobenius norm is written as $\| \cdot \|_F$. We denote the $p\times p$ identity matrix by $I_p$ and the null matrix by $0_{p}$. Positive definite and semidefinite matrices are indicated via $A \succ 0$ and $A \succeq 0$, respectively. As we will be dealing with block matrices and vectors, subscripts $A_{ij}$ are used to indicate matrix blocks and $V_i$, vector blocks, respectively. In the case of iterates, we  use parentheses as $A_{(k)}$ to indicate the $k$-th iteration.

Throughout this paper, we will be considering Euclidean transformations of $\mathbb{R}^p$. More concretely, targeting applications where $p=3$. The orthogonal group can be realized as the matrix group
\begin{equation}
    \mathrm{O}(p) = \big\{R \in \mathbb{R}^{p\times p} \;|\; R^\top R = RR^\top = I\big\},
\end{equation}
and the special orthogonal group as the connected component of $\mathrm{O}(p)$ with positive determinant
\begin{equation}
    \mathrm{SO}(p) = \big\{R \in \mathbb{R}^{p\times p} \;|\; R \in \mathrm{O}(p),\;\det(R)=1\big\}.
\end{equation}
The Special Euclidean group is defined as
\begin{equation}
    \mathrm{SE}(p) = \big\{t \in \mathbb{R}^p, R \in \mathrm{SO}(p) \;|\; (t,R)\big\}.
\end{equation}
We borrow some linear operators defined in \cite{Rosen2019}, that vastly simplify the notation, namely $\mathrm{Diag}$ and $\mathrm{BlockDiag}$. Given a $p\times p$-block matrix $U \in \mathbb{R}^{np\times np}$, $\mathrm{BlockDiag}$ extracts the block diagonal
\begin{equation}
    \mathrm{BlockDiag}(U) = \begin{bmatrix}
    U_{11} & \dots & 0 \\
    \vdots & \ddots & \vdots \\
    0 & \dots & U_{nn}
    \end{bmatrix}.
\end{equation}
Conversely, $\mathrm{Diag}$, forms a $np\times np$ block diagonal matrix from a set of $n$ blocks of dimension $p\times p$
\begin{equation}
    \mathrm{Diag}(U_1,\dots,U_n) = \begin{bmatrix}
    U_{1} & \dots & 0 \\
    \vdots & \ddots & \vdots \\
    0 & \dots & U_{n}
    \end{bmatrix}.
\end{equation}

Finally, let us recall some elements of graph theory. We refer to graphs as $G = (\mathcal{V},\mathcal{E})$, with $\mathcal{V}=\{1,\dots,|\mathcal{V}|\}$ the set of vertices and $\mathcal{E}=\{e_1,\dots,e_{|\mathcal{E}|}\}\subseteq \mathcal{V} \times \mathcal{V}$ the set of edges. Here, $|\cdot |$ indicates the cardinality of the respective set. While we identify each graph vertex with an integer, every edge $e_k \in \mathcal{E}$ is identified with a tuple consisting of its incident vertices. Thus, the edge from vertex $i$ to $j$ will henceforth be indicated by the tuple $(i,j)$. The set of neighbors of a vertex $i$ is denoted as $\delta_i$. The degree of a vertex $i \in V$ is the number of its neighbors and it is therefore written as $\vert \delta_i \vert$. A weighted graph is a graph wherein we have a real-valued function $w$ defined on the set of edges as $w_{ij}$. By comparison, in an unweighted graph we assume this function to be identically 1. From the set of edge weights we can define the graph's adjacency matrix $A\in\mathbb{R}^{\vert\mathcal{V}\vert \times \vert\mathcal{V}\vert}$ as
\begin{equation}
    A_{ij} := \begin{cases}
             w_{ij}, \quad & (i,j) \in \mathcal{E} \\
             0,\quad & \mathrm{otherwise}
             \end{cases}
\end{equation}
and the graph Laplacian as $L := \mathrm{Diag}(\delta_1,\dots,\delta_{\vert \mathcal{V}\vert}) - A$.

\section{Related Work}
Research on PGO has undergone a remarkable evolution. In this section, we hope to shed light on a set of the works that have contributed towards it. Keeping in line with the theme of this article, we will focus on the Maximum Likelihood Estimation (MLE) formulation of synchronization problem on $\mathrm{SO}(p)$. For a review of robust synchronization, the reader is referred to \cite{sidhartha2021all,purkait2020neurora,lee2022hara}. 

MLE is one of the most widely adopted approaches to PGO. Under the circumstances typically encountered in robotics and computer vision applications, the isotropic Gaussian and Langevin \cite{Boumal2014} noise models adopted therein provide an accurate description of real-world phenomena. That being the case, what may be gained in accuracy through this approach, gives way to reduced tractability due to the non-convexity of the problem, associated with $\mathrm{SO}(p)$. In fact, unrestricted retrieval of optimal solutions in this group is only possible for cycle graphs \cite{Peters2015,Moreira2021}. 

For this reason, earlier works all shared one common ground: that of being local search methods, with no optimality guarantees. Such was the case of Gauss-Newton based-techniques, as the ones implemented in the popular frameworks g2o \cite{Kummerle2011} and GTSAM \cite{Dellaert2012}. These initialization-dependent iterative approaches, ultimately led to a large body of works concerning \textit{good initializations} i.e., initial pose guesses from where the aforementioned methods are expected to attain the optimum.  These initializations, often heuristic in nature, solve a simpler proxy problem in $\mathrm{SO}(p)$ and may be iterative themselves e.g., the graph consensus formulation by Tron et al. \cite{Tron2008} which is solved on the $\mathrm{SO}(p)$ manifold directly using the Riemannian gradient. Alternatively, the closed-form spectral initialization featured in \cite{Arrigoni2016,Rosen2019,Boumal2016}, which involves the projection of the $p$ eigenvectors of the connection Laplacian corresponding to the smallest eigenvalues, yields a good approximation of the optimum for moderate noise levels. This approximation is well understood at this point, with suboptimality bounds given for $\mathrm{SO}(2)$ \cite{Boumal2016} and $\mathrm{SO}(3)$ \cite{doherty2022performance}. In addition, it lends itself to high-performance implementations by leveraging the sparsity of the graph \cite{Moreira2021wacv}. For a more complete survey and benchmark of PGO initializations, see Carlone et al. \cite{Carlone2015}.

While the combination of Gauss-Newton with an appropriate initialization makes for a good strategy to avoid suboptimal accumulation points, the issue of optimality verification persists. This was ultimately overcome via duality theory. While, as we mentioned, problem (\ref{p:general_pgo}) is non-convex, its dual and bi-dual are SDPs which can be solved in polynomial time via interior-point methods, the latter being the convex relaxation of (\ref{p:general_pgo}). This convexification of the problem has been the go-to strategy in recent research since it provides either an optimality certificate or a means to ascertain how suboptimal a given solution is. Eriksson et al. \cite{Eriksson2018} derived the dual and bi-dual SDP of the problem in $\mathrm{SO}(p)$ i.e., rotation averaging, and solve the latter using a block-coordinate descent approach. Carlone et al. \cite{Carlone2015_dual} also rely on the dual to solve the primal in $\mathrm{SE}(3)$. For phase synchronization in the complex circle group, Boumal \cite{Boumal2016} proposed the Generalized Power Method, which is an application of the Frank-Wolfe method, achieving linear convergence \cite{Liu2017}. 

Apart from the non-convexity, the second difficulty inherent to PGO is the high-dimensionality of the problem. If on the one hand the bi-dual SDP allow us to bypass PGO's non-convex nature, on the other hand, the new problem involves a $np\times np$ dense matrix variable, which in common applications can easily reach upwards of $10^8$ entries. To address this issue, Rosen et al. \cite{Rosen2019} leveraged the results of Burer-Monteiro \cite{Burer2003} on the low-rank solutions of SDPs and proposed a Riemannian staircase \cite{Boumal2015_riemann} to solve the SDP via increasing rank factorizations. In the same paper, the authors also derive bounds for the tightness of the convex relaxation i.e., the duality gap, as a function of the graph connectivity and the degradation level of the measurements. This effectively proved that in well connected graphs there is greater leeway in how degraded the pairwise measurements can be, before solutions are no longer certifiable. This low-rank approach to the SDP was also featured in Shonan averaging \cite{Dellaert2020}, wherein the rank increase is realized by searching over SO(q) for $q \geq p$.

\section{Problem statement}
\label{sec:problem statement}
% Indicar que a contribuição vem na sec 5
We start by the noise model commonly adopted in the literature and the nonconvex MLE problem derived upon it. We then set forth the Lagrangian and the SDP dual problem.

Consider a set of $n$ vertices $\mathcal{V}=\{1,\dots,n\}$. We parameterize each vertex with a latent position $\underline{t}_i \in \mathbb{R}^p$ and a latent orientation $\underline{R}_i \in \mathrm{SO}(p)$. In other words, each vertex $i\in \mathcal{V}$ is associated with a pose $\underline{x}_i\in \mathrm{SE}(p)$ i.e., a rigid (Euclidean) transformation $(\underline{R}_{i}, \underline{t}_{i})$. For every pairwise tuple $ (i,j) \in \mathcal{V}^2$, we define relative transformations as $x_{ij} := x_i x_j^{-1}$. It follows that $\underline{R}_{ij} := \underline{R}_i \underline{R}_j^\top$ encodes the latent relative orientation of vertex $i$ w.r.t. vertex $j$ and $\underline{t}_{ij} = \underline{t}_i - \underline{R}_i^\top \underline{t}_j$ the position of the $i$-th vertex as seen from the reference frame of vertex $j$. The input to a PGO problem is a set of noisy relative pose measurements $\{(\widetilde{R}_{ij},\tilde{t}_{ij})\}$, for $(i,j) \in \mathcal{E}\subseteq \mathcal{V}^2$. The goal is to estimate the $n$ latent poses $\{(\underline{R}_i, \underline{t}_i)\}_{i\in \mathcal{V}}$ that originated those relative measurements. 

We assume that $G$ is connected. This hypothesis is necessary due to an inherent ambiguity of this estimation problem. Without anchoring any of the vertices, the set of poses $\{(R_i,t_i)\}_{i\in\mathcal{V}}$ is equivalent to $\{R_i \hat{R}, R_i \hat{t} + t_i\}_{i\in \mathcal{V}}$ for any rigid transformation $(\hat{R},\hat{t})$. This is known as gauge freedom. In Fig. \ref{fig:translations_pgo} e.g., we can globally rotate and translate the entire mesh, without affecting the relative poses of the vertices. If the underlying graph is not connected, the disconnected components may each be independently optimized up to a rigid transformation, but never with respect to a common reference frame, which is undesirable. We also assume that the graph $G$ is cyclic, i.e., it has at least one cycle, since it is the existence of cycles that leads to inconsistencies and thus to an optimization problem. In the absence of cycles, the only possible pose estimates are given by the composition of relative transformations along paths originating in a common vertex.

\subsection{MLE formulation of PGO}
The model most commonly adopted in the literature consists of additive isotropic white Gaussian noise for the translation variables
\begin{equation}
    \widetilde{t}_{ij} = \underline{t}_{ij} + e_{ij} = \underline{t}_i - \underline{R}_j^\top\underline{t}_j + e_{ij},
\end{equation}
where $e_{ij} \sim \mathcal{N}(0,\tau^{-1}_{ij} I_d)$ with pdf 
\begin{equation}
    F_{\text{Gaussian}}(e; 0, \tau^{-1} I_d) \propto \exp\bigg(-\frac{\tau}{2}\|e\|_2^2\bigg)
\end{equation}
and isotropic Langevin noise for the rotations
\begin{equation}
    \widetilde{R}_{ij} = \underline{R}_{ij}R_{ij}^\epsilon = \underline{R}_i \underline{R}_j^\top R_{ij}^\epsilon,
\end{equation}
where $R_{ij}^\epsilon \sim \text{Langevin}(I_p,\kappa_{ij})$, with $\kappa_{ij}$ being the concentration parameter with the Langevin pdf \cite{Boumal_langevin}
\begin{equation}
    F_{\text{Lang}}(R^\epsilon;I_p,\kappa) \propto \exp\big(\kappa \Tr(R^\epsilon{}^\top)\big).
\end{equation}
Under the assumption of measurement independence, the log-likelihood function reads as
\begin{align}
    f_\text{ML} = \sum_{i\sim j} \kappa_{ij} \| R_i - &  \widetilde{R}_{ij} R_j \|_F^2 -\tau_{ij}\| \widetilde{t}_{ij} - \underline{t}_i + \underline{R}_j^\top\underline{t}_j \|_2^2.
    \label{eq:logfml}
\end{align}
The log-likelihood $f_\text{ML}$ is non-concave. However, it is concave in the translation variables. In \cite{Rosen2019}, the authors showed that maximizing the log-likelihood is equivalent to solving first a rotation-only synchronization problem and then computing the optimal translations, which is a least-squares problem. We can thus focus our attention on the problem of finding the optimal rotations. Let the connection adjacency $\tilde{A} \in \mathbb{R}^{np\times np}$ for the rotational problem be defined as the symmetric $p\times p$-block matrix
\begin{equation}
    \tilde{A}_{ij} := \begin{cases}
            \kappa_{ij}\tilde{R}_{ij},\quad & (i,j) \in E \\
            0_{p\times p}, \quad & \mathrm{otherwise}.
    \end{cases}
\end{equation}
Letting $d_i := \sum_{i\sim j}\kappa_{ij}$ and $D \in \mathbb{R}^{np\times np}$ the block diagonal
\begin{equation}
    D:=\mathrm{Diag}\big(d_1 I_p,\dots,d_n I_p\big),
    \label{eq:the_D}
\end{equation}
we have the connection Laplacian given by $\tilde{L} := D-\tilde{A}$. Stacking the rotation matrices horizontally in a column block-vector $R \in \mathrm{SO}(p)^n \subset \mathbb{R}^{np\times p}$
\begin{equation}
    R := \begin{bmatrix}
    R_1^\top & \dots & R_n^\top
    \end{bmatrix}^\top,
\end{equation}
the rotation synchronization problem can be stated as the minimization of a quadratic in $\mathrm{SO}(p)^n$,
\begin{equation}
    \begin{aligned}
        &\underset{R}{\text{minimize}} & & -\big\langle R R^\top, \tilde{A} - \tilde{Q}^\tau \big\rangle \\
        &\text{subject to} & & R \in \mathrm{SO}(p)^n,
    \end{aligned}
\label{p:pgo_problem}
\end{equation}
where $\tilde{Q}^\tau$ is a dense matrix containing the translation terms \cite{Rosen2019}. If we only retain the $\tilde{A}$ term, we obtain the rotation averaging problem \cite{hartley2013rotation,Eriksson2018,Dellaert2020}
\begin{equation}
    \begin{aligned}
        &\underset{R}{\text{minimize}} & & -\big\langle R R^\top, \tilde{A} \big\rangle \\
        &\text{subject to} & & R \in \mathrm{SO}(p)^n.
    \end{aligned}
\label{p:rotation_averaging_problem}
\end{equation}
Problems (\ref{p:pgo_problem}) and (\ref{p:rotation_averaging_problem}) are both instances of the orthogonal-cut problem. The former providing the solution for measurements in $\mathrm{SE}(p)$ and the latter used either as a rotation-only initialization for the former or as a standalone problem for measurements in $\mathrm{SO}(p)$, with applications such as clock synchronization in $\mathrm{SO}(2)$ and structure-from-motion in $\mathrm{SO}(3)$. In what follows, we will focus on the Rotation Averaging Problem (\ref{p:rotation_averaging_problem}), considering the concentration parameters $\kappa_{ij}$ to be identically 1. 

\subsection{From non-convex MLE to strong duality}
% - Explicar o porquê desta seccao
We shall now briefly summarize the theory underlying optimality certification for Problem (\ref{p:rotation_averaging_problem}). We refer the reader to \cite{Eriksson2018, Rosen2019} for the detailed proofs. Let $\Lambda := \mathrm{Diag}(\Lambda_1,\dots,\Lambda_n)$ be a symmetric $p\times p$-block diagonal dual variable. The Lagrangian of Problem (\ref{p:rotation_averaging_problem}) is
\begin{equation}
    \mathcal{L}(R,\Lambda) = -\big\langle R, \tilde{A} R \big\rangle - \big\langle\Lambda, (R R^\top - I_{np})\big\rangle.
\end{equation}
Differentiating $\mathcal{L}$ w.r.t. $R$, we have the first-order KKT condition
\begin{equation}
    \tilde{A} R = \Lambda R,
    \label{eq:kkt}
\end{equation}
which allows us to retrieve $\Lambda$ from $\tilde{A} R$ as $\Lambda_i = \sum_{j}\tilde{A}_{ij}R_j R_i^\top$ and conversely, $R$ from the kernel of $\Lambda-\tilde{A}$. The dual of Problem ($\ref{p:rotation_averaging_problem}$) is the SDP
\begin{equation}
    \begin{aligned}
        &\underset{\Lambda}{\text{maximize}} & & -\Tr(\Lambda) \\
        &\text{subject to} & & \Lambda - \tilde{A} \succeq 0.
    \end{aligned}
\label{p:rotation_averaging_dual}
\end{equation}
From duality theory\cite{Eriksson2018,Rosen2019}, if $\exists\; R, \Lambda$ feasible such that (\ref{eq:kkt}) holds, then $\Lambda - \tilde{A} \succeq 0$, is enough to guarantee that the pair $(R, \Lambda)$ is primal-dual optimal i.e., the duality gap between the primal (\ref{p:rotation_averaging_problem}) and the dual (\ref{p:rotation_averaging_dual}) is zero. The matrix $\Lambda - \tilde{A}$ provides thus an optimality certificate, provided it is positive-semidefinite. In fact, let $\lambda_1 \leq \dots \leq \lambda_{np}$ be the eigenvalues of $\Lambda - \tilde{A}$. Then, for $R, \Lambda$ primal-dual feasible, the duality gap is bounded below as
\begin{align}
    -\big\langle R, \tilde{A} R \big\rangle + \Tr\big(\Lambda\big) \geq np(\lambda_1 + \lambda_2 + \lambda_3).
    \label{eq:dual_gap_eigenvalues}
\end{align}
For connected graphs, the latent connection Laplacian defined as $\underline{L} := D - \underline{A}$ is both positive semidefinite and has nullity of $p$. Thus, not only is $D$ the best candidate for the optimal $\Lambda^\ast$, but in this case the optimum $R^\ast$, which is equal to its latent counterpart $\underline{R}$, actually lies in the kernel of the $\underline{L}$. Evidently, in the real world the latent connection Laplacian is unknown. In addition, the real connection Laplacian $\tilde{L}=D-\tilde{A}$ no longer has a kernel.
Nevertheless, $D$ which is known from the graph topology (the graph degree matrix), is still dual-feasible and may be a good approximation of $\Lambda^\ast$. The suboptimality of this estimate is, according to (\ref{eq:dual_gap_eigenvalues}), bounded by the smallest eigenvalues of $\tilde{L}=D-\tilde{A}$. This is the crux of the spectral initialization, which takes the eigenspace spanned by the $p$ eigenvectors of $\tilde{L}$ corresponding to the smallest eigenvalues, and projects them orthogonally to $\mathrm{SO}(p)^n$. As we shall expound in the subsequent section, the proposed primal-dual method exploits this further by starting with $\Lambda_{(0)}=D$ and iteratively updating the dual estimate.

\section{Primal-Dual Method}
\label{sec:primal dual}
% dizer que gpm é novo/diff do iccv

In this section, we present a primal-dual method to solve Problem (\ref{p:rotation_averaging_problem}). The building blocks to arrive at a primal-dual method are the first-order KKT condition (\ref{eq:kkt}) of the problem and the Generalized Power Method (GPM) proposed in \cite{Boumal2016}. The former will be used to compute primal updates $R_{(k)}$, in a similar fashion to the spectral initialization. The latter is used to produce the dual estimates $\Lambda_{(k)}$.

\subsection{Primal update}
Given an estimate of the dual variable e.g., $D$, the KKT condition (\ref{eq:kkt}) will in general not have a solution in $\mathrm{SO}(p)^n$. The spectral initialization resorts thus to an approximation, by first relaxing the orthogonality constraint $\mathrm{BlockDiag(X X^\top)}=I_{np}$ for $X^\top X = n I_p$, i.e. it supplants orthogonality of the $p\times p$ blocks of $X$ for column orthogonality. The relaxation thereby obtained
\begin{equation}
    \argmin_{X^\top X = n I_p} \big\langle X X^\top , (D - \tilde{A}) \big\rangle,
\end{equation}
can be solved by computing the eigenvectors of $D - \tilde{A}$ corresponding to the smallest eigenvalues. These eigenvectors are then projected to $\text{SO}(p)^n$ via an orthogonal projection, to obtain the feasible primal estimate, denoted by $R_{(0)}$. 

The focal point of our update rule lies in starting with the spectral initialization i.e., $\Lambda_{(0)}=D$, but updating this block diagonal matrix afterwards with a better, yet infeasible, dual estimate $\Lambda_{(k)}$. For the primal update, given $\Lambda_{(k)}$ at the $k$-th iteration, we solve thus
\begin{align}
    X_{(k+1)} &= \argmin_{X^\top X = n I_p} \big\langle X X^\top , (\Lambda_{(k)} - \tilde{A})\big\rangle \label{eq:problem_in_stiefel} \\
    R_{(k+1)} &= \argmin_{R \in \mathrm{SO}(p)^n} \big\vert\big\vert R - X_{(k+1)} \big\vert\big\vert_F^2 . \label{eq:procrustes}
\end{align}

The gauge-invariant distance from this estimate to the global optimum $R^\ast$ of Problem (\ref{p:rotation_averaging_problem}), which we write as 
% bounds no fim da seccao
\begin{equation}
    d_{(0)}:=\min_{G \in \mathrm{SO}(p)} \|R^\ast - R_{(0)} G \|_F,
\end{equation}
was bounded by Doherty et al. \cite{doherty2022performance} for $k=0$ as
\begin{equation}
    d_{(0)} \leq \frac{(8+4\sqrt{2})\sqrt{np} \| \tilde{A}-\underline{A}\|_2}{\lambda_{p+1}(\underline{L})},
    \label{eq:doherty_bound}
\end{equation}
wherein we witness the influence of the noise in $\| \tilde{A}-\underline{A}\|_2$ and the inverse relationship with the connectivity of the graph, indicated by its Fiedler value $\lambda_{p+1}(\underline{L})$. In our case, the bound in (\ref{eq:doherty_bound}) becomes
\begin{align}
    d_{(k)}
    \leq (8+4\sqrt{2})\sqrt{np}\frac{\|\big(\Lambda_{(k)}-D\big)- (\tilde{A}-\underline{A})\|_2}{\lambda_{p+1}(\underline{L})}.
\end{align}

Similarly to the spectral initialization, the solution of (\ref{eq:problem_in_stiefel}) can be computed efficiently by means of sparse Krylov-based symmetric eigensolvers, as in \cite{Moreira2021wacv}. The optimization problem in (\ref{eq:procrustes}) can be solved via $n$ SVDs of $p\times p$ matrices. Further, given a primal-dual pair $(R^\ast,\Lambda^\ast)$ that verifies the stationarity condition (\ref{eq:kkt}), $\Lambda^\ast - \tilde{A} \succeq 0$ provides an optimality certificate. Thus, (\ref{eq:problem_in_stiefel}) allows for an optimality assessment at each iteration.

\subsection{Dual update}
In order to formulate the dual update, we start by recalling the recursion of GPM. Define the orthogonal projection from $\mathbb{R}^{np\times p}$ to $\mathrm{SO}(p)^n \subset \mathbb{R}^{np\times p}$ as the map 
\begin{align}
    P (X) = \argmin_{R \in \mathrm{SO}(p)^n} \| X- R \|_F^2,
    \label{eq:ortho_proj}
\end{align}
which has a closed-form solution given by the blockwise SVD of of $X$. GPM is realized via fixed-point iterations of the operator $P$ i.e., 
\begin{equation}
    R_{(k+1)} = P\Big(\tilde{A} R_{(k)} \Big).
    \label{eq:gpm}
\end{equation}
As noted by the authors, this is an application of the Frank-Wolfe method since $\tilde{A} R_{(k)}$ is the linearization of the quadratic in Problem (\ref{p:rotation_averaging_problem}). Alternatively, it may also be viewed as a projected gradient method. Convergence of (\ref{eq:gpm}) is shown to be linear in $\mathrm{SO}(2)$ \cite{Liu2017}. 

We will now write the fixed-point iterations (\ref{eq:gpm}) in terms of the dual variable, thus reformulating GPM itself as an infeasible primal-dual method. Compute the SVD of the $i$-th block $\big(\tilde{A} R_{(k)}\big)_i$ as $U_i\Sigma_i V_i^\top$ and set
\begin{equation}
    \Lambda_{(k)} = \mathrm{Diag}\big(U_1\Sigma_1 U_1^\top,\dots,U_n\Sigma_n U_n^\top\big).
\end{equation}
Then, the GPM update may be written as
\begin{equation}
    R_{(k+1)} = {\Lambda_{(k)}}^{-1} \tilde{A} R_{(k)}.
    \label{eq:dual_gpm}
\end{equation}
If $R_{(\infty)}$ and $\Lambda_{(\infty)}$ are accumulation points of the sequence defined in (\ref{eq:dual_gpm}) then, $R_{(\infty)} = \Lambda_{(\infty)}^{-1} \tilde{A} R_{(\infty)}$, which is the KKT condition (\ref{eq:kkt}) of the primal problem. Further, similarly to the monotonous increase in $\langle R_{(k)}, \tilde{A} R_{(k)}\rangle$, we have for $\Lambda_{(k)}$ the following result.
\begin{lemma}
\label{lemma:monotonous_lambda}
The dual iterates $\Lambda_{(k)}$ of GPM defined by the recursion (\ref{eq:dual_gpm}) verify $\Tr(\Lambda_{(k+1)}) \geq \Tr(\Lambda_{(k)})$ (proof in the appendix).
\end{lemma}

Note that, in spite of this reformulation of GPM as a primal-dual method, the iterates $\Lambda_{(k)}$ are not dual-feasible. 
\begin{lemma}
\label{lemma:dual_infeasible}
Assume $\tilde{A} \succeq 0$. The updates defined by the recursion (\ref{eq:dual_gpm}) are dual infeasible (proof in the appendix). 
\end{lemma}

To form the dual update, we rely on the reformulation of GPM just presented, in particular the recursion in (\ref{eq:dual_gpm}). Given a primal estimate $R_{(k)}$, we compute an iteration of GPM as
\begin{equation}
    \tilde{A} R_{(k)} = \Lambda_{(k)} Y.
\end{equation}
Here, $Y\in\mathrm{SO}(p)^{n}\subset \mathbb{R}^{np\times p}$ would be the next primal estimate in the GPM sequence. However, we only retain the dual estimate $\Lambda_{(k)}$. In more explicit notation, from the SVDs
\begin{equation}
    \sum_{j\in\delta_i} \tilde{A}_{ij} R_{{(k)}_j} = U_i \Sigma_i V_i^\top,\quad i=1,\dots,n
\end{equation}
we set $\Lambda_{(k)_i} = U_i \Sigma_i U_i^\top$. Note that this approach is comparable to the work of Gao et al. \cite{gao2019} on optimization problems with orthogonality constraints, wherein the dual update is achieved by symmetrizing $\sum_{j \in \delta_i} \tilde{A}_{ij} R_{{(k)}_j} R_{{(k)}_i}^\top$ i.e., 
\begin{align}
    \Lambda_{(k)_i} = \frac{1}{2}\sum_{j\in\delta_i}\tilde{A}_{ij} {R_{(k)}}_{j} R_i^\top + \frac{1}{2}\bigg(\sum_{j\in\delta_i}\tilde{A}_{ij} {R_{(k)}}_{j} R_i^\top \bigg)^\top
\end{align}
In Algorithm \ref{algo:rotavg} we show how our primal-dual updates were implemented. The parameter $\sigma$ used in the sparse eigensolver corresponds to the eigenvalue target for the eigenvectors we are computing. Since our primal update is achieved by solving (\ref{eq:problem_in_stiefel}), we pick $\sigma < 0$ such that the three eigenvectors retrieved correspond to the three smallest eigenvalues. Prior to projecting the solution of the primal problem we fix the gauge freedom by anchoring the first rotation, chosen arbitrarily.

\begin{algorithm}
\SetAlgoLined
\SetKwInOut{Input}{input}
\SetKwInOut{Output}{output}
\Input{$\tilde{A}, D$ \tcp*[f] {Adjacency and degree}}
\Output{$R,\Lambda$}
$\Lambda \gets D$

\For {$t\gets 0$ to $\text{maxiter}$}{
    $X, \lambda_1, \dots, \lambda_p \gets \text{eigensolver}\big(\Lambda - \tilde{A})$
    $X \gets X X_1^{-1}$ \tcp*[f] {Gauge fix}
    
    \For {$i\gets 2$ to $n$}{
        $U \Sigma V^\top \gets \text{SVD}\big(X_i\big)$
        $R_i \gets U \;\text{diag}\big([1\; 1\; \dots \; \text{det}(UV^\top)]\big)\; V^\top$
    }
    $Y \gets \tilde{A} R$ \\
    \For {$i\gets 1$ to $n$}{
        $U_i \Sigma_i V_i^\top \gets \text{SVD}(Y_i)$ \\ 
        $\Lambda_i \gets U_i \Sigma_i U_i^\top $
    }
    $\Lambda \gets \text{BlockDiag}(\Lambda_1,\dots,\Lambda_n)$\;
    \If{$\min\{\vert\lambda_1\vert, \dots, \vert\lambda_p\vert\} < \epsilon$}{
        \Return $R, \Lambda$
    }
}
\caption{Primal-Dual in SO(p)}
\label{algo:rotavg}
\end{algorithm}

We conclude this section by noting that the primal-dual iterations can easily be adapted to the problem of angular synchronization, by using the complex circle group $S=\{z\in \mathbb{C} \;|\; \vert z \vert = 1\}$ to represent rotations, instead of the matrix group $\mathrm{SO}(2)$. In this case, $\tilde{A} \in \mathbb{C}^{n\times n}$ is Hermitian, with $\tilde{A}_{ij} = \tilde{z}_{ij}$ for $(i,j)\in \mathcal{E}$. In the spectral step of the primal update, only one eigenvector is computed, corresponding to the smallest eigenvalue. This eigenvector is element-wise projected to $S$ via $z_i \gets z_i/\vert z_i \vert$. For the dual update, $\Lambda$ is a real $n\times n$ diagonal matrix. Instead of the SVD, $\Lambda_i$ is updated to the magnitude of $\sum_{j\in\delta_i} \tilde{z}_{ij} z_j$.

\section{Cycle graphs}
\label{sec:cycle_graphs}
From a topological standpoint, cycle graphs are the simplest instance of rotation averaging. Implementation-wise, they are the ones where sparsity may be exploited the most. From a theoretical viewpoint, closed-form solutions for the global optima are known \cite{Peters2015}. A natural consequence of this is that they present an adequate benchmark for iterative methods \cite{Eriksson2018, Dellaert2020}. In this section, we characterize the solutions of the stationary points for rotation averaging problems with an underlying cycle graph topology, global optima included. We do so in via a different approach than the one adopted in \cite{Peters2015}. We will first derive the closed-form solutions for one-parameter subgroups of $\mathrm{SO}(3)$. We then show that, in the general case, there is a basis wherein the connection adjacency matrix $\tilde{A}$ can be written such that the measurements lie in a one-parameter subgroup of $\mathrm{SO}(3)$. These results culminate in the proof that the primal-dual iterations (Algorithm \ref{algo:rotavg}) retrieve the global optimum in cycle graphs, after one iteration.

We start by defining the error $E \in \mathrm{SO}(3)$ incurred while traversing the cycle graph starting and ending on the same node. Without loss of generality let
\begin{equation}
    E := \prod_{k=0}^{n-1} \widetilde{R}_{\mathrm{mod}(k,n)+1,\;\mathrm{mod}(k+1,n)+1},
    \label{eq:def_cycle_error}
\end{equation}
with the matrix product in (\ref{eq:def_cycle_error}) being defined from left to right. In a cycle with 3 nodes e.g., $E = \widetilde{R}_{12}\widetilde{R}_{23}\widetilde{R}_{31}$. 

Further, let $\gamma \in [-\pi,\pi]$ be the angle of $E$, which we denote by $\gamma := \angle(E)$. We define the set of the $n$-th roots of $E$ as
\begin{equation}
    E^{\frac{1}{n}} := \big\{E_0,E_1,\dots,E_{n-1}\big\},
\end{equation}
with $E_k \in \mathrm{SO}(3)$, $E_k^n = E$ and $\angle(E_k) = \gamma/n - 2k\pi/n$, for  $k \in \{0,\dots,n-1\}$.

\subsection{One-parameter subgroups of SO(3)}
\label{sec:optimization_in_so2}
We consider for now one-parameter subgroups of $\mathrm{SO}(3)$ by assuming that the pairwise rotation measurements $\widetilde{R}_{12},\widetilde{R}_{23},\dots,\widetilde{R}_{n1}$ share a common axis. We start by characterizing its stationary points.

\begin{lemma} For cycle graphs whose edge measurements lie in a one-parameter subgroup of $\mathrm{SO}(3)$, the points
\begin{equation}
    R_i = \Bigg(\prod_{s=1}^{i-1} \widetilde{R}_{s,s+1}\Bigg)^\top E_k^{i-1},\quad i\in\{2,\dots,n\}
    \label{eq:stationary_solution_cycle_so2}
\end{equation}
with $R_1 = I_3$, indexed by $k\in\{0,\dots,n-1\}$, are stationary points of problem (\ref{p:rotation_averaging_problem}).
\label{prop:solution_in_so2}
\end{lemma}
From Lemma \ref{prop:solution_in_so2}, we arrive at the optimal solution in SO(2) in the following result.
\begin{theorem}
For cycle graphs whose edge measurements lie in a one-parameter subgroup of $\mathrm{SO}(3)$, the point
\begin{equation}
    R_i^\ast = \Bigg(\prod_{k=1}^{i-1} \widetilde{R}_{k,k+1}\Bigg)^\top E_0^{i-1},\quad i\in\{2,\dots,n\}
    \label{eq:optimal_solution_cycle_so2}
\end{equation}
with $R_1^\ast = I_3$ is a solution of problem (\ref{p:rotation_averaging_problem}).
\begin{proof}
Invoking Theorem 4.2 of \cite{Eriksson2018} for cycle graphs, strong duality will hold and a solution will be globally optimal if $\forall i\sim j$ the residuals verify $\big\vert \widetilde{\theta}_{ij} - \theta^\ast_{ij} \big\vert \leq \frac{\pi}{n}$. From (\ref{eq:optimal_solution_cycle_so2}), the optimal rotations verify
\begin{equation}
    R^\ast_{j} =  E_0\widetilde{R}_{ij}^\top R^\ast_{i},
    \label{eq:recursion_optimal}
\end{equation}
with $\angle(E_0)=\gamma/n$. From (\ref{eq:recursion_optimal}) we can write
\begin{equation}
    \widetilde{\theta}_{ij} - \theta^\ast_{ij} = \angle\big(\widetilde{R}_{ij} R_{j}^\ast R_{i}^{\ast\top}\big) = \gamma/n,
\end{equation}
Since
\begin{equation}
    \vert \gamma /n \vert \leq \frac{\pi}{n}
\end{equation}
due to $\gamma \in [-\pi, \pi]$, the solution in (\ref{eq:optimal_solution_cycle_so2}) is optimal.
\end{proof}
\label{prop:global_min_so2}
\end{theorem}
In cycle graphs, rotation averaging problems in one-parameter subgroups of $\mathrm{SO}(3)$ will redistribute the cycle error equitably over all of the edges. If we incur a cycle error of $E$, with $\angle(E)=\gamma$, the optimal relative rotation $R_i^\ast R_j^{\ast\top} $ will have an angular residual of $\gamma/n$ relative to the respective measurement $\widetilde{R}_{ij}$. By increasing this figure by a multiple of $2\pi / n$ we obtain suboptimal stationary points of Problem (\ref{p:rotation_averaging_problem}). Borrowing the spring system analogy of Section \ref{sec:introduction}, in stationary points the net force  at each vertex will be zero. In a cycle, this can only happen if the spring displacements i.e., the residuals, are the same everywhere.

%% -- Number of equation is not (73)

\subsection{Optimization in SO(3)}
We now show that any cycle graph problem in $\mathrm{SO}(3)$ has the same expression for its stationary points (\ref{eq:stationary_solution_cycle_so2}) and global optimum (\ref{eq:optimal_solution_cycle_so2}) as derived for one-parameter subgroups. We accomplish this by rewriting $\tilde{A}$ in a new basis. 

Define the matrix $U \in \mathrm{SO}(3n)$ as
\begin{equation}
     U := \mathrm{BlockDiag}\big(U_1,\dots,U_n\big),
     \label{eq:defu}
\end{equation}
with $U_i \in \mathrm{SO}(3)$ for $i \in \{1,\dots,n\}$ computed according to
\begin{equation}
U_i :=
\begin{cases}
    I_3, \quad & i=1 \\
    \widetilde{R}_{i-1, i}^\top U_{i-1}, \quad & i\in\{2,\dots,n\}.
    \label{eq:defu2}
\end{cases}
\end{equation}

Denote by $\tilde{A}'$, the matrix $\tilde{A}$ written in the basis $U$ i.e., $\tilde{A}' := U^\top \tilde{A} U$. From (\ref{eq:defu2}), the blocks on the lower triangular part of $\tilde{A}':=U^\top \tilde{A}U$ are given by
\begin{equation}
    \tilde{A}'_{ij} = 
    \begin{cases}
            U_n^\top \widetilde{R}_{n1} U_1, & i=n, j=1 \\
            U_i^\top \widetilde{R}_{ij} \widetilde{R}_{ij}^\top U_i, & j=i-1 \\
            0_{3\times 3}, & i=j.
    \end{cases}
    \label{eq:blocks_in_bases}
\end{equation}
It is immediate that for $j=i-1$ we have $\widetilde{R}'_{ij}=I_3$. It suffices to show that $U_n^\top \widetilde{R}_{n1} U_1 = E$. Note that from (\ref{eq:defu2}) we have $U_1=I_3$ and $U_n = \widetilde{R}_{n-1,n}^\top \dots \widetilde{R}_{23}^\top \widetilde{R}_{12}^\top $. Therefore,
\begin{equation}
    U_n^\top \widetilde{R}_{n1} U_1  = \widetilde{R}_{12}\widetilde{R}_{23}\dots \widetilde{R}_{n-1,n}\widetilde{R}_{n1},
    \label{eq:un_replaced}
\end{equation}
which equals $E$ i.e., the cycle error (\ref{eq:def_cycle_error}), by definition of the latter. Thus,
\begin{align}
    \tilde{A}' = \begin{bmatrix}
    0_{3\times 3} & I_3 & \dots & 0 & E^\top \\
    I_3 & 0_{3\times 3} & \dots & 0 & 0 \\
    \vdots & \vdots & \ddots & \vdots & \vdots\\
    0 & 0 & \dots & 0_{3\times 3} & I_3 \\
    E & 0 & \dots & I_3 & 0_{3\times 3} \\
    \end{bmatrix}
    \label{eq:Rtildeprime}
\end{align}

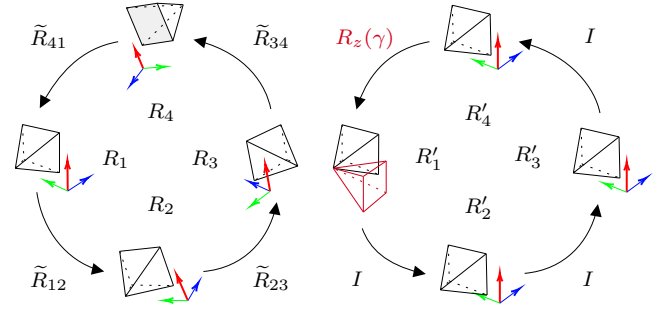
\begin{figure}
\centering
\input{equivalentcycles.tikz}
\vspace{-0.3cm}
\caption{The cycle graph problem on the left can be transformed into the problem on the right via a change-of-basis.}
\label{fig:equivalence}
\end{figure}

We can visualize this result in Fig. \ref{fig:equivalence}. In cycle graphs, MLE rotation averaging can be solved by concentrating the cycle error $E$ at a single edge. Further, by changing basis, the pairwise measurements $I_p$ and $E$ belong to a one-parameter subgroup of $\mathrm{SO}(3)$. We can thus leverage the results from Section \ref{sec:optimization_in_so2} to retrieve the global optimum and stationary points of problem in closed-form.

\begin{theorem} For cycle graphs with edge measurements in $\mathrm{SO}(3)$, the point
\begin{align}
    {R}_i^\ast &= \Bigg(\prod_{s=1}^{i-1} {\widetilde{R}}_{s,s+1} \Bigg)^\top E_0 ^{i-1},\quad i\in\{2,\dots,n\}
\end{align}
with $R_1^\ast= I_3$, is a solution of the problem (\ref{p:rotation_averaging_problem}).
\label{prop:optimal_in_so3}

\begin{proof}
We rewrite the cost function as as
\begin{equation}
    f(R) = \big\langle\big(U^\top R,  U^\top \tilde{A} U (U^\top R)\big\rangle.
\end{equation}
Using the change-of-variables $R' = U^\top R$ and the change-of-basis $\tilde{A}' = U^\top \tilde{A} U$ we have the equivalent problem
\begin{equation}
    \begin{aligned}
        &\underset{R'}{\textrm{minimize}} & & \big\langle {R'}, \tilde{A}' R'\big\rangle \\
        &\textrm{subject to} & &  R' \in \mathrm{SO}(3)^n,
    \end{aligned}
    \label{eq:newbasisproblem}
\end{equation}
whose edge measurements either $I_p$ or $E$. These rotations belong to the one-parameter subgroup $t\mapsto \exp\big(t[\hat{n}]_\times\big)$, where $\hat{n}$ is the axis of $E$. Theorem \ref{prop:global_min_so2} is thus applicable and the solution of (\ref{eq:newbasisproblem}) is
\begin{equation}
    {R'}_i^\ast =  E_0^{i-1}, \quad i\in\{1,\dots,n\},
    \label{eq:solution_in_so2}
\end{equation}
since $\widetilde{R}'_{i,i+1}=I_3, \; \forall i\in\{1,\dots,n-1\}$. It suffices now to write (\ref{eq:solution_in_so2}) in the old basis vectors according to $R^\ast=U {R'}^\ast$. Since $U$ is block-diagonal, $R_i^\ast = U_i {R'}_i^\ast$. From the definition of $U_i$ (\ref{eq:defu2}) we have
\begin{equation}
    {R}_i^\ast = \Bigg(\prod_{s=1}^{i-1} {\widetilde{R}}_{s,s+1} \Bigg)^\top E_0 ^{i-1},\quad i\in\{2,\dots,n\}
\end{equation}
with ${R}_1^\ast = I_3$.
\end{proof} 
\end{theorem}
As a corollary of Theorem \ref{prop:optimal_in_so3}, we can take any stationary point of the problem in the new basis (see Lemma \ref{prop:solution_in_so2}) and revert to the old basis vectors in order to obtain the corresponding stationary point of problem. Thus, the points
\vspace{-0.3cm}
\begin{equation}
    R_i = \Bigg(\prod_{s=1}^{i-1} {\widetilde{R}}_{s,s+1} \Bigg)^\top E_k ^{i-1},\quad i\in\{2,\dots,n\}
    \label{eq:evalf}
\end{equation}
with $R_1=I_3$, are stationary points of (\ref{p:rotation_averaging_problem}) indexed by $k\in\{0,\dots,n-1\}$, where the cost function evaluates to
\begin{equation}
    f(R) = -2n \Tr\big(E_k\big).
    \label{eq:stationary_law}
\end{equation}
Since $\Tr\big(E_k\big)=1+2\cos(\gamma/n-2k\pi/n)$ it follows that the greater the number of nodes, the greater the number of local minima near the global optimum. Hence the difficulty of solving rotation averaging optimally. 

Further, in cycle graphs the spectrum of $\tilde{A}$ relates to the values of the cost function at stationary points and can therefore be computed in closed-form.
\begin{theorem}
\label{theo:spectrum}
Let $\sigma\big(\tilde{A}\big)$ denote the spectrum of $\tilde{A}$. Then,
\begin{align}
    \sigma\big(\tilde{A}\big) = &\big\{2\cos\big(\gamma/n - 2k\pi/n\big) \big\}_{k=0,\dots, n-1}\nonumber \\
    &\cup \big\{2\cos\big(2k\pi/n\big)\big\}_{k=0,\dots, n-1}
\end{align}
(proof in the appendix).
\end{theorem}

We conclude this section by showing that, for cycle graphs the primal-dual method (Algorithm \ref{algo:rotavg}) retrieves the rotation averaging global optimum in $\mathrm{SO}(3)$ at the first iteration. Letting $\Lambda_{(0)}=D$, from the proof of Theorem \ref{theo:spectrum}, we know that the equation for the  eigenvectors of $\tilde{A}'$ corresponding to the three largest eigenvalues is
\begin{equation}
    \tilde{A}' \begin{bmatrix}
    I \\ E \\ E^2 \\ \vdots \\ E^{n-1}
    \end{bmatrix} J = \begin{bmatrix}
    I \\ E \\ E^2 \\ \vdots \\ E^{n-1}
    \end{bmatrix} J \Sigma.
    \label{eq:blocks_eigenvector}
\end{equation}
Changing bases, $\tilde{A}' = U^\top \tilde{A} U$ and attending to the definition of $U$ (\ref{eq:defu}) yields
\begin{equation}
    \tilde{A} R^\ast J = R^\ast J \Sigma.
    \label{eq:blocks_eigenvector}
\end{equation}
Thus, the first spectral update of the primal-dual method satisfies the equation
\begin{equation}
    \big(\Lambda_{(0)} - \tilde{A}\big)R^\ast J = R^\ast J (2I_3-\Sigma).
\end{equation}
Since $R^\ast J  \in \mathrm{SO}(3)^n$, the orthogonal projection is the identity. The dual update is then given via the blockwise SVD of $\tilde{A} R^\ast J$ i.e.,
\begin{equation}
    \Lambda_{(1)} = \mathrm{BlockDiag}\big(R_1^\ast J \Sigma J^\top {R_1^\ast}^\top,\dots,R_n^\ast J \Sigma J^\top {R_n^\ast}^\top \big)
\end{equation}
which verifies $\big(\Lambda_{(1)}-\tilde{A}\big)R^\ast = 0$.

\section{Experimental results}
In this section, we evaluate the performance of our primal-dual method and our closed-form solution in pose graph datasets and synthetic rotation averaging problems in cycle graphs, respectively. Our algorithms were implemented in C++ and all the tests were conducted on a laptop computer with a 6-core Intel Core i7-9750H@2.6GHz.

\subsection{Optimality}
\label{section:optimality}
\begin{figure}
    \centering
    \includegraphics[width=\linewidth,trim={1cm 8cm 1.5cm 8.5cm},clip]{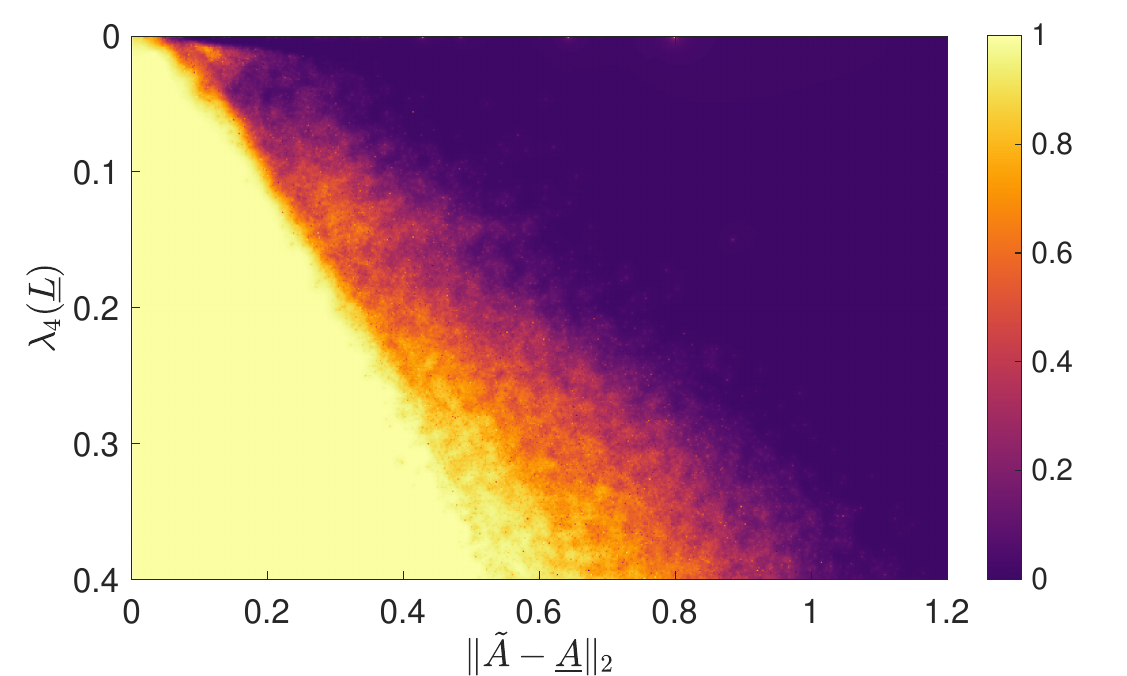}
    \caption{Primal-dual convergence to the optimum in SO(3) (color represents frequency). Vertical axis: Graph's Fiedler value (same as the 4th eigenvalue of the latent connection Laplacian $\underline{L}$). Horizontal axis: operator norm of the difference between the latent and measured connection adjacency matrices, $\underline{A}$ and $\tilde{A}$, respectively. Axes in multiples of $n$.}
    \label{fig:convergence}
\end{figure}

In order to analyze the conditions under which the primal-dual iterations proposed succeeded in retrieving the global optimum, we will invoke one key result from \cite{Rosen2019}, Proposition 2 on the exact recovery of the solution via the SDP relaxation. This proposition states that $\exists \;\beta > 0$, function of the ground-truth connection Laplacian $\underline{L}$ such that if $\|\tilde{L} - \underline{L}\|_2 < \beta$, the duality gap is zero. This spectral norm is an indication of how degraded the measurements are. Since we expect $\beta$ to depend on the graph connectivity, in Fig. \ref{fig:convergence} we plot the frequency of optimal solution retrieval for $10^5$ combinations of the underlying graph's Fiedler value (vertical axis) and the aforementioned spectral norm (horizontal axis). By inspecting the plot, we can see that for larger graph connectivities, a larger maximum noise threshold below which certifiable optimal recovery is possible. The relationship between the two is approximately linear.

\begin{figure*}
\centering
\subfloat{\label{4figs-a} \includegraphics[width=0.245\textwidth,trim=1.5cm 0 2cm 0,clip]{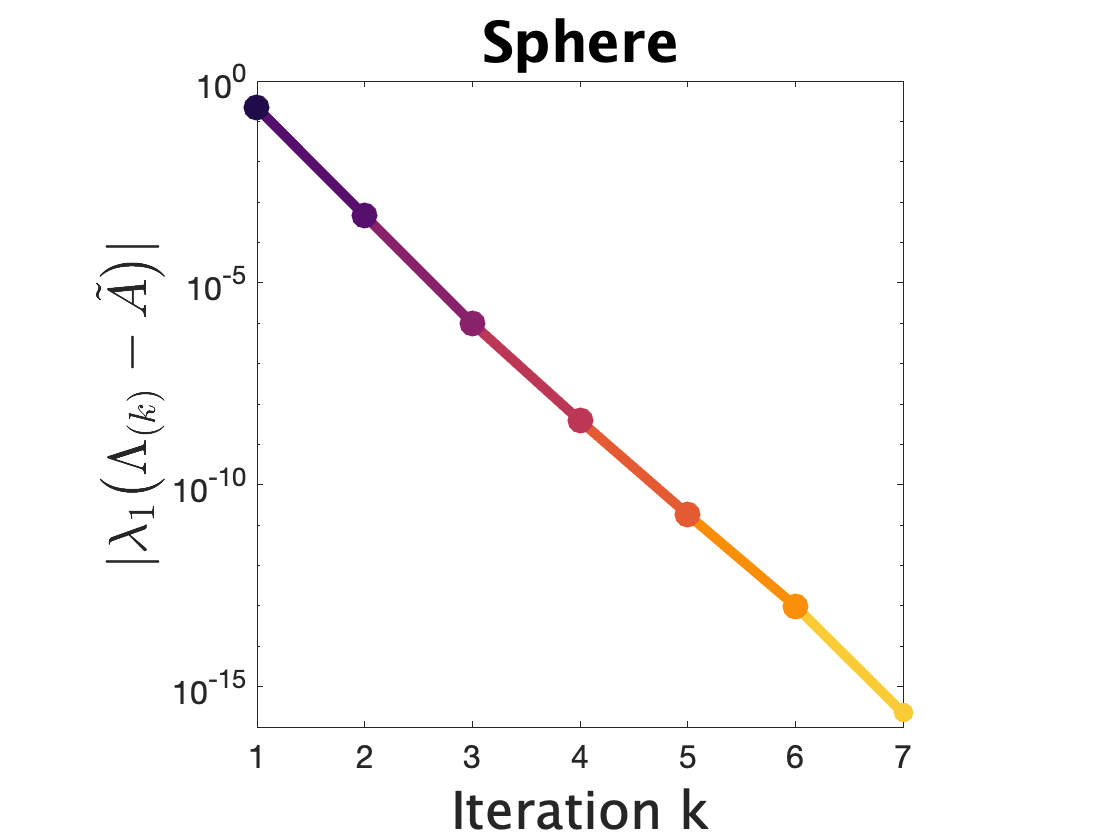}}% The "%" masks the line break.
\hfill
\subfloat{\label{4figs-b} \includegraphics[width=0.245\textwidth,trim={1.5cm 0 2cm 0},clip]{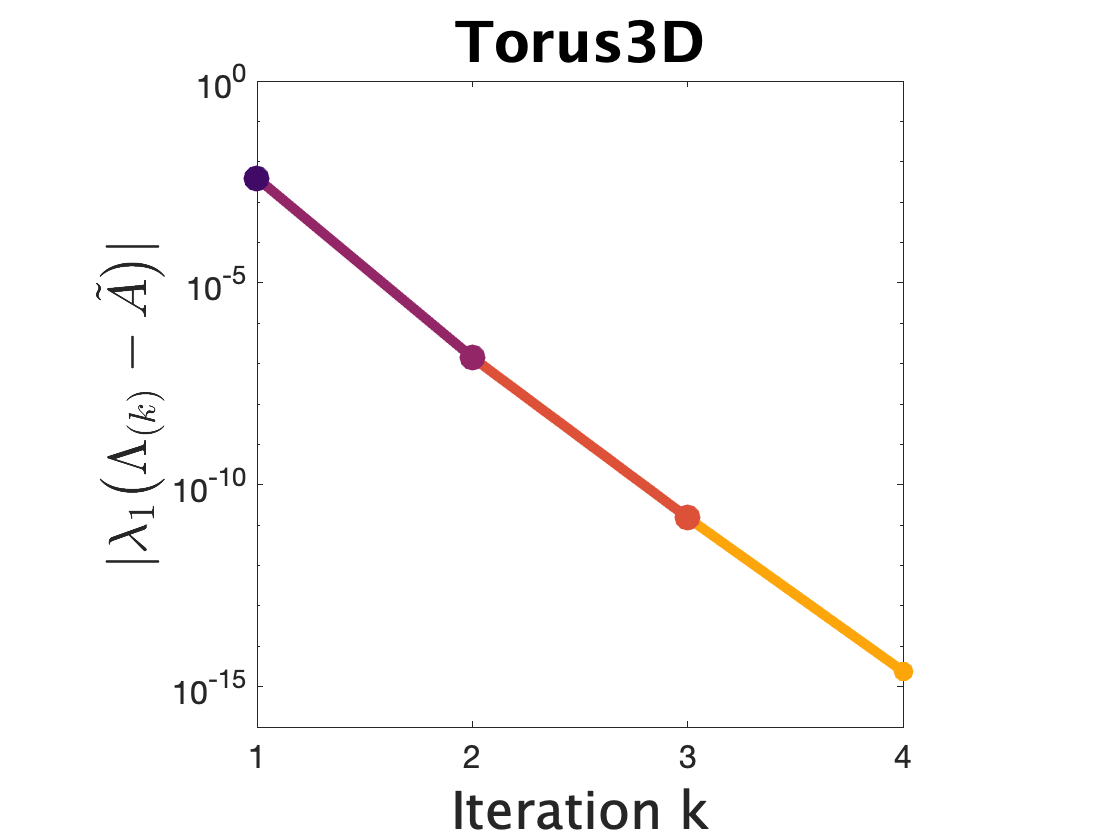}}%
\hfill
\subfloat{\label{4figs-c} \includegraphics[width=0.245\textwidth,trim={1.5cm 0 2cm 0},clip]{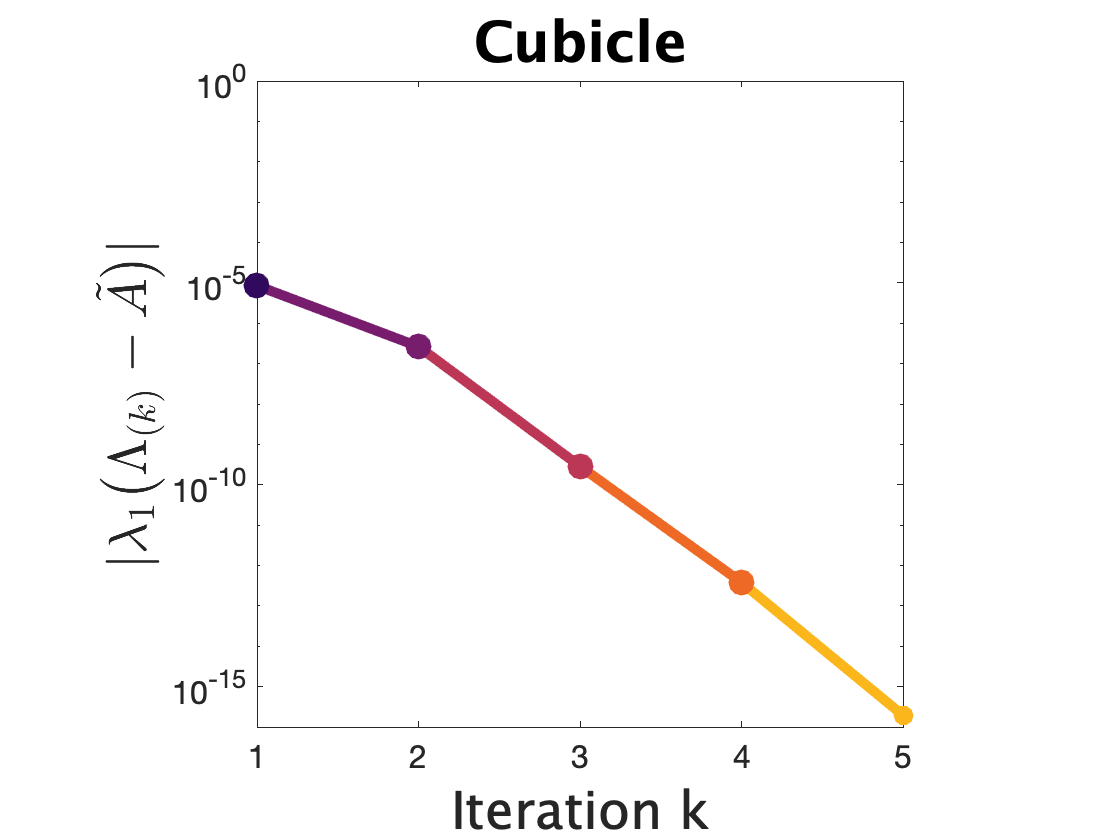}}%
\hfill
\subfloat{\label{4figs-d} \includegraphics[width=0.245\textwidth,trim={1.5cm 0 2cm 0},clip]{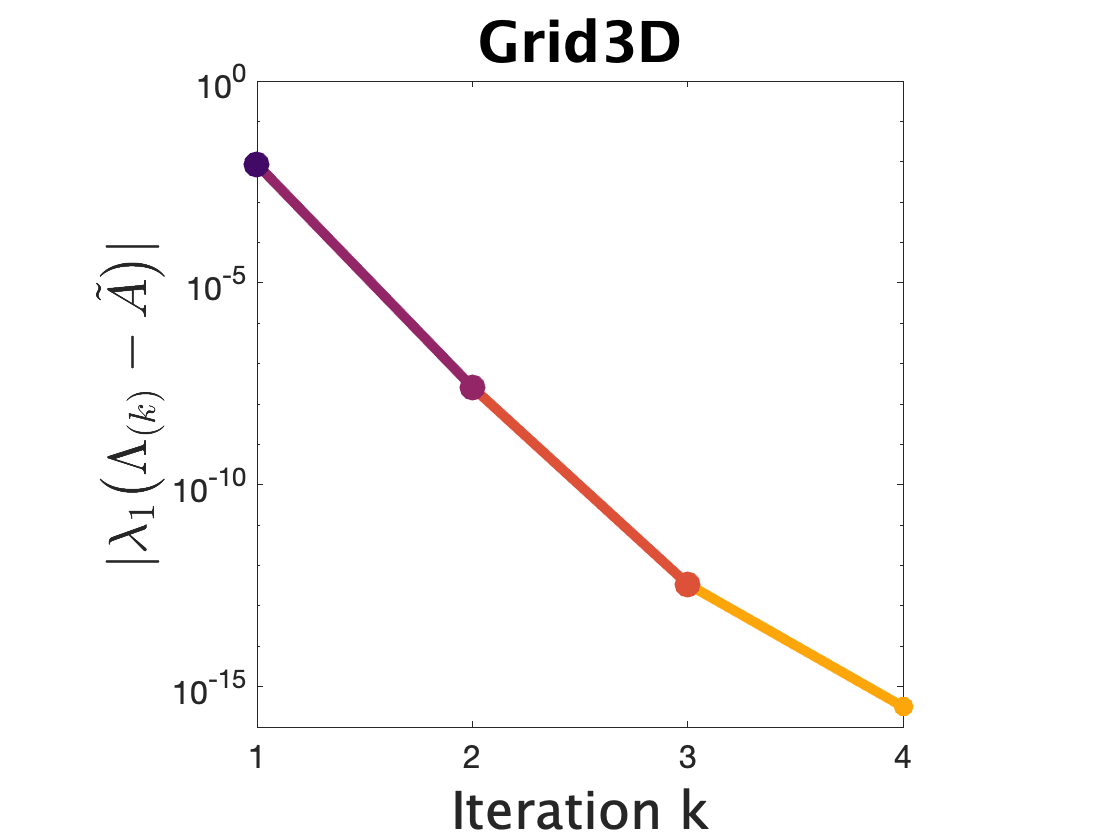}}%
\caption{Convergence of absolute value of the smallest eigenvalue of $\Lambda_{(k)}-\tilde{A}$ for the PGO datasets Sphere (0.36s), Torus3D (0.35s), Cubicle (0.46s) and Grid3D (1.78s), made available in \cite{Carlone2015_dual, Carlone2015}.}
\label{4figs}
\end{figure*}

\subsection{Cycle graphs}
Borrowing the evaluation approach adopted in \cite{Dellaert2020, Eriksson2018}, we tested our primal-dual algorithm, in synthetic cycle graph datasets. These consisted of random rotation averaging problems with underlying cycle graph structures of different sizes, wherein the ground-truth absolute orientations $R_i$ correspond to rotations around the z-axis, forming a circular trajectory. The synthetic pairwise measurements were simulated by perturbing the relative ground-truth orientations between adjacent nodes by an error matrix obtained from angle-axis representations. The axes were sampled uniformly over the unit sphere. The angles were drawn from a normal distribution with zero mean and standard deviation $\sigma$. 

\begin{figure}
    \centering
    \includegraphics[width=\linewidth, trim={0.9cm 2.0cm 0.6cm 1.5cm},clip]{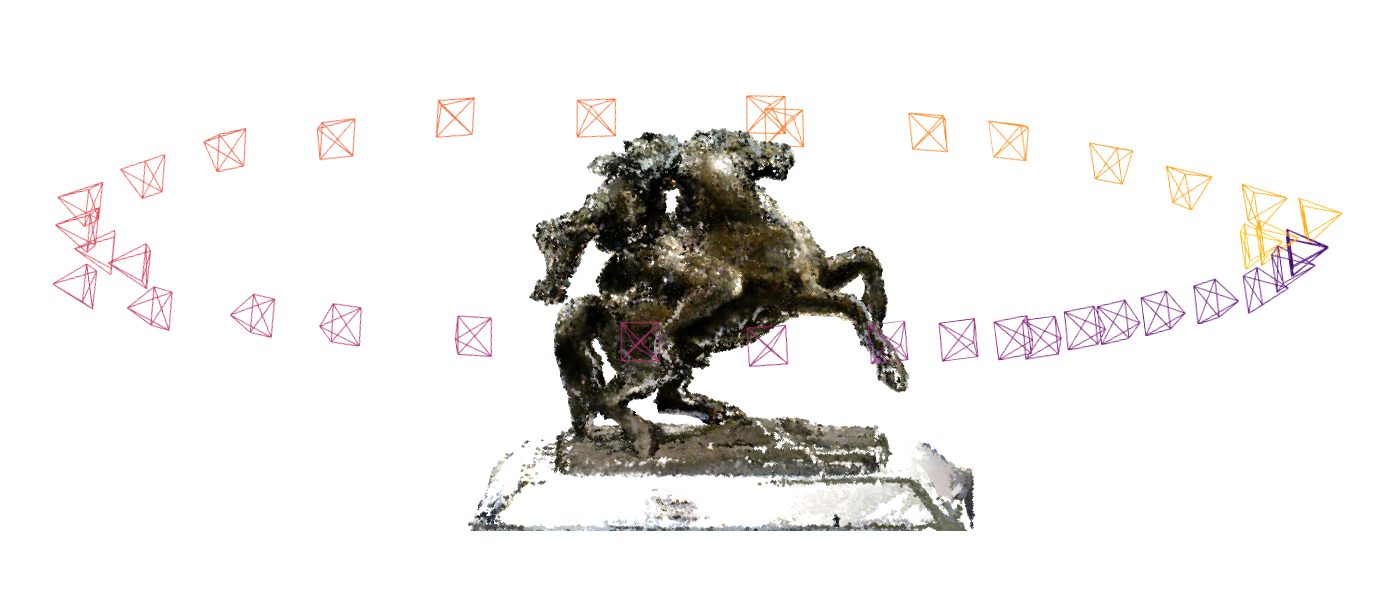}
    \caption{Cycle graph 3D reconstruction with 38 poses. Rotation averaging using the primal-dual algorithm (\ref{algo:rotavg}) takes 2ms. For reference, our implementation of the closed-form solution takes 0.07ms.}
    \label{fig:horse}
\end{figure}

We benchmarked our method against two baselines, the block coordinate descent method (BCD) \cite{Eriksson2018} used to solve the bi-dual of problem (\ref{p:rotation_averaging_problem}) and the SA algorithm which we also tested in our pose graph experiments. We implemented the former in MATLAB and used the author's implementation of the latter. Both methods were initialized randomly. Results averaged over 20 simulations can be observed in Table \ref{tab:cycle_benchmark}. For our solution, we list the average of the smallest eigenvalue of $\Lambda^\ast - \tilde{A}$, denoted by $\vert\bar{\lambda}_1\vert$, which certifies that, as we have shown, our solution is optimal to machine precision in all the simulations we ran. In the two rightmost columns, we show the average difference $\bar{\Delta}^\ast$ between our closed-form global minimum and the cost function evaluated at the set of rotations produced by SA and BCD.

Using its default settings, SA retrieved the global optimum in all the tests conducted. Nevertheless, not only does precision wane as the order of the cycle increases but also the average CPU time surges substantially as the number of variables increases. In order to test the BCD method, we first computed the global optimum in each simulation via the closed-form solution. We then used it to set the stopping criterion for the BCD as $\Delta^\ast \leq 10^{-3}$. As evidenced by the average CPU time, convergence flatlined for the largest cycles. Attaining the global minimum to three decimal places using this algorithm took, on average, as much as 96 seconds for $n=100$ and longer than that would be required for $n = 200$. While this may be a shortcoming of our implementation, the orders of magnitude of the CPU time appear to be in accordance with those reported in \cite{Dellaert2020}.

\subsection{PGO benchmarks}
Using seven datasets from the pose graph optimization literature available online \cite{Carlone2015}, we extracted the pairwise rotation measurements from each one in order to generate rotation averaging problems. Some of these datasets contain multiple measurements per edge, from which only one was kept. We compare the performance the primal-dual method in Algorithm \ref{algo:rotavg}) against Shonan Averaging (SA) \cite{Dellaert2020} and SE-Sync \cite{Rosen2019}. The authors' implementations are available online and we tested them with their default parameters. Since SE-Sync is designed for solving pose graph optimization problems, we set the input translations to zero. The stopping criterion for our method was defined as $\vert\lambda_1\vert < 10^{-15}$, which corresponds to tolerance of the Krylov-based eigensolver used.

The results can be observed in Table \ref{tab:slam3d_benchmark}. In order to juxtapose the three methods in terms of the positive semidefiniteness of $\Lambda - \tilde{A}$ i.e., in order to verify optimality, we proceeded as follows. For each estimate computed, we obtained the Lagrange multiplier using the KKT condition in (\ref{eq:kkt}) and symmetrized it. The columns $\lambda_1$ in Table \ref{tab:slam3d_benchmark} correspond to the minimum eigenvalue of $\Lambda - \tilde{A}$, which is zero if a given solution is optimal and strong duality holds. In addition, we represent the cost function evaluated at the solution produced by Algorithm \ref{algo:rotavg}, denoted by $f^\ast$ and the difference between this minimum and the minima computed by SA and SE-Sync, denoted by $\Delta^\ast = f^\ast - f$. The CPU time is shown in seconds for all three algorithms.

The three methods benchmarked reach the global optimum in all seven datasets. While there may be disparities regarding precision, the differences in terms of the minimum attained and the set of rotations produced are negligeable in the applications considered. We focus thus on the CPU time of each algorithm. Our primal-dual method attains machine precision of $\lambda_1$, and therefore the global optimum, faster than the two other methods take to stop iterating. If we were to relax the upper bound on our stopping criterion, the CPU times could be brought down even further, without compromising the solution as far as geometric reconstructions are concerned. Plots showcasing the convergence of the primal-dual method, in terms of the positive-semidefinitess of $\Lambda-\tilde{A}$, for four of the datasets are shown in Fig. \ref{4figs}.

\begin{table*}[h]
\small
\caption{Comparison between primal-dual iterations (Algorithm \ref{algo:rotavg}), Shonan Averaging \cite{Dellaert2012} and SE-Sync \cite{Rosen2019}. Datasets from \cite{Carlone2015}.}
\begin{center}
\setlength{\tabcolsep}{5.5pt} % Default value: 6pt
\renewcommand{\arraystretch}{1.1}
\begin{tabular}{|l|cc|crc|ccr|ccr|}
\hline
& \multicolumn{2}{c|}{Graph} & \multicolumn{3}{c|}{Primal-dual} & \multicolumn{3}{c|}{Shonan averaging} &  \multicolumn{3}{c|}{SE-Sync} \\
Dataset & $n$  & $m$ & $\vert\lambda_1\vert$ & $f^\ast $ & $t (s)$ &  $\vert\lambda_1\vert$ & $\Delta^\ast$ (approx.) & $t (s)$  & $\vert\lambda_1\vert$ & $\Delta^\ast$ (approx.) & $t (s)$ \\
\hline\hline
SmallGrid  & $125$ & $297$ & $\mathbf{10^{-15}}$ & $\mathbf{-2118.202}$ & $\mathbf{0.02}$ & $10^{-07}$ & $-10^{-04}$ & 0.55 & $10^{-08}$ &   $-10^{-05}$  & 0.09 \\
Garage    & $1661$ & $6275$ & $\mathbf{10^{-15}}$ & $\mathbf{-42632.998}$  & $\mathbf{0.06}$  & $10^{-05}$ & $-10^{-01}$ & 24.5 & $10^{-14}$ & $-10^{-10}$ & 0.99\\
Sphere    & $2200$ & $8647$ & $\mathbf{10^{-15}}$ & $\mathbf{-56981.692}$ & $\mathbf{0.36}$ & $10^{-07}$ & $-10^{-03}$ & 30.1  & $10^{-09}$ & $-10^{-05}$ & 2.79 \\
Torus3D   & $5000$ & $9048$ & $\mathbf{10^{-15}}$ & $\mathbf{-69227.058}$ & $\mathbf{0.35}$ & $10^{-06}$ & $-10^{-02}$ & 98.8 & $10^{-09}$ & $-10^{-05}$ & 3.86 \\	
Cubicle   & $5750$ & $12486$ &  $\mathbf{10^{-15}}$ & $\mathbf{-92163.079}$ & $\mathbf{0.46}$ & $10^{-05}$ &  $-10^{-02}$ & 96.8 & $10^{-08}$  & $-10^{-04}$ & 2.49 \\
Grid3D    & $8000$ & $22236$ & $\mathbf{10^{-15}}$ &  $\mathbf{-157206.257}$ & $\mathbf{1.78}$ & $10^{-07}$ &  $-10^{-02}$ & 154.54 & $10^{-09}$  & $-10^{-04}$ & 11.69 \\
Rim       & $10195$ & $22251$ &  $\mathbf{10^{-15}}$ &  $\mathbf{-164037.930}$ & $\mathbf{5.92}$  & $10^{-05}$  & $-10^{-01}$ &  221.63 & $10^{-12}$ & $-10^{-07}$ & 8.73\\
\hline
\end{tabular}
\end{center}
\label{tab:slam3d_benchmark}
\end{table*}

\begin{table}
\small
\caption{Comparison between Primal-dual, Shonan Averaging \cite{Dellaert2020} and the block coordinate descent method (BCD) \cite{Eriksson2018} for random cycle graph problems.}
\begin{center}
\setlength{\tabcolsep}{5.5pt} % Default value: 6pt
\renewcommand{\arraystretch}{1.1}
\begin{tabular}{|cc|cr|cr|cr|}
\hline
\multicolumn{2}{|c|}{Problem} & \multicolumn{2}{c|}{Primal-dual} & \multicolumn{2}{c|}{Shonan} &  \multicolumn{2}{c|}{BCD} \\
$n$  & $\sigma$ (rad) & $\vert \lambda_1 \vert$ & $\bar{t}$ (s) & $\bar{\Delta}^\ast$ & $\bar{t}$ (s) & $\bar{\Delta}^\ast$ & $\bar{t}$  (s)\\
\hline\hline
$20$ & $0.2$ & $\mathbf{10^{-15}}$  &  \textbf{0.019} & $10^{-4}$ & 0.11 & $10^{-3}$  & 0.18 \\
 & $0.5$ & $\mathbf{10^{-15}}$  &  \textbf{0.019} &  $10^{-4}$ & 0.12 & $10^{-3}$ &  0.23 \\

$50$ & $0.2$ & $\mathbf{10^{-15}}$  &  \textbf{0.020} & $10^{-3}$ & 0.26 & $10^{-3}$ &  4.48 \\
 & $0.5$ & $\mathbf{10^{-15}}$  &  \textbf{0.020} & $10^{-2}$ & 0.32 & $10^{-3}$ &   6.80 \\

$100$ & $0.2$ & $\mathbf{10^{-15}}$  & \textbf{0.022} &  $10^{-2}$ & 0.42 & $10^{-3}$ & 51.75 \\
 & $0.5$ & $\mathbf{10^{-15}}$  & \textbf{0.022}  & $10^{-2}$ & 0.50 & $10^{-3}$  &  96.19 \\

$200$ & $0.2$ & $\mathbf{10^{-15}}$  & \textbf{0.024} & $10^{-2}$ &  0.74 & $10^{-3}$  & n.a.  \\
 & $0.5$ & $\mathbf{10^{-15}}$  & \textbf{0.024} & $10^{-1}$  &  1.10 & $10^{-3}$  & n.a. \\
\hline
\end{tabular}
\end{center}
\label{tab:cycle_benchmark}
\end{table}

\section{Conclusion}
In this paper, we addressed the problem of MLE rotation averaging, as an initialization for PGO in SE(3), or as a standalone problem. We set forth a primal-dual method with spectral updates, and a characterization of stationary points for problems with an underlying cycle graph topology. Our proposed algorithm extends the well-known spectral initialization by updating the dual variable and, as demonstrated by our empirical evaluation, it produces optimal solutions to machine precision in a fraction of the time of existing solvers. Further, it verifies optimality by default at each iteration. We expect this work to open up new avenues of research, namely the extension of the primal-dual iterations to the complete PGO problem in SE(3) and new in terms of new theoretical results regarding the convergence properties of the proposed method.

% if have a single appendix:
%\appendix[Proof of the Zonklar Equations]
% or
%\appendix  % for no appendix heading
% do not use \section anymore after \appendix, only \section*
% is possibly needed

% use appendices with more than one appendix
% then use \section to start each appendix
% you must declare a \section before using any
% \subsection or using \label (\appendices by itself
% starts a section numbered zero.)
% you can choose not to have a title for an appendix
% if you want by leaving the argument blank

\appendices
\section*{Proof of Lemma \ref{lemma:monotonous_lambda}}
From the update $\Lambda_{(k)} R_{(k+1)} = \tilde{A} R_{(k)}$ write
\begin{equation}
R_{(k+1)}^\top \Lambda_{(k)} R_{(k+1)} = R_{(k+1)}^\top \tilde{A} R_{(k)}
\end{equation}
Taking the trace on both sides 
\begin{equation}
\Tr\big(\Lambda_{(k)}\big) = \big\langle R_{(k+1)},  \tilde{A} R_{(k)} \big\rangle
\end{equation}
From $ \tilde{A} R_{(k+1)} =  \Lambda_{(k+1)}R_{(k+2)}$
\begin{equation}
\Tr\big(\Lambda_{(k)}\big) = \big\langle R_{(k+2)}, \Lambda_{(k+1)}  R_{(k)} \big\rangle \leq \Tr\big(\Lambda_{(k+1)}\big) 
\end{equation}
where the inequality arises from the general fact that $\Tr(SR)\leq\Tr(S)$ for $S$ symmetric and $R$ orthogonal.

\section*{Proof of Lemma \ref{lemma:dual_infeasible}}
Start by noting that $\Lambda_{(k)} - \tilde{A} \succeq 0$ is equivalent to $I_{np} - \Lambda_{(k)}^{-\frac{1}{2}} \tilde{A} \Lambda_{(k)}^{-\frac{1}{2}} \succeq 0$ and write the GPM iteration as
\begin{equation}
    \Lambda^\frac{1}{2}_{(k)} R_{(k+1)} = \Big( \Lambda^{-\frac{1}{2}}_{(k)} \tilde{A}  \Lambda^{-\frac{1}{2}}_{(k)} \Big) \Lambda^\frac{1}{2}_{(k)} R_{(k)}
\end{equation}
Using $\| \Lambda^\frac{1}{2}_{(k)} R_{(k+1)} \|_2 = \| \Lambda^\frac{1}{2}_{(k)} R_{(k)} \|_2$, the submultiplicative norm property yields
\begin{equation}
    \Big\| \Lambda^{-\frac{1}{2}}_{(k)} A \Lambda^{-\frac{1}{2}}_{(k)}  \Big\|_2 \geq 1
\end{equation}
Since $A \succeq 0 \implies \Lambda^{-\frac{1}{2}}_{(k)} \tilde{A}  \Lambda^{-\frac{1}{2}}_{(k)} \succeq 0$, we have 
\begin{equation}
     \Big\| \Lambda^{-\frac{1}{2}}_{(k)} \tilde{A} \Lambda^{-\frac{1}{2}}_{(k)}  \Big\|_2 = \lambda_{np}\Big(\Lambda^{-\frac{1}{2}}_{(k)} \tilde{A}  \Lambda^{-\frac{1}{2}}_{(k)}\Big)
\end{equation}
Thus,
\begin{align}
\lambda_{1}\Big(I_{np} - \Lambda^{-\frac{1}{2}}_{(k)} & \tilde{A} \Lambda^{-\frac{1}{2}}_{(k)}\Big) \nonumber \\ 
&= 1 - \lambda_{np}\Big(\Lambda^{-\frac{1}{2}}_{(k)} \tilde{A} \Lambda^{-\frac{1}{2}}_{(k)}\Big) \leq 0
\end{align}
Since the smallest eigenvalue of $I_{np} - \Lambda^{-\frac{1}{2}} \tilde{A} \Lambda^{-\frac{1}{2}}$ is always non greater than zero, the iterations must comprehend infeasible dual estimates.

\section*{Proof of Lemma \ref{prop:solution_in_so2}}
We rewrite the cost function in (\ref{p:rotation_averaging_problem}) as
\begin{equation}
    f(R)= -2\sum_{i\sim j}\Tr\big(\widetilde{R}_{ij}R_jR_i^\top\big).
    \label{eq:trace}
\end{equation}
Under the hypothesis that the rotations $\widetilde{R}_{ij}$ share a common axis, we can restrict our search for $R_1,\dots,R_n$ to this subgroup. Thus, $\angle(\widetilde{R}_{ij}R_jR_i^\top)=\angle(\widetilde{R}_{ij}) - \angle(R_i R_j^\top)$. From $\Tr(R) = 1 + 2\cos(\angle(R))$, the trace in (\ref{eq:trace}) becomes
\begin{align}
    \Tr\big(\widetilde{R}_{ij}R_jR_i^\top\big) = 1+2\cos\big(\angle(\widetilde{R}_{ij}) - \angle(R_i R_j^\top)\big).
\end{align}
Define the angles $\theta_{ij}:=\angle(R_i R_j^\top)$, $\widetilde{\theta}_{ij}:=\angle(\widetilde{R}_{ij})$ and the set $\Theta=\{\theta_{12},\dots,\theta_{n1}\}$. The optimization problems
\begin{equation}
    \begin{aligned}
        &\underset{\Theta}{\textrm{maximize}} & & \sum_{i \sim j} \cos\big(\widetilde{\theta}_{ij} - \theta_{ij}\big) \\
        &\textrm{subject to} & &  \theta_{12} + \dots + \theta_{n1} = 2k\pi,
    \end{aligned}
    \label{eq:cosine_problem}
\end{equation}
for $k\in\{0,\dots,n-1\}$, yield stationary points of (\ref{p:rotation_averaging_problem}). Let the residuals be $\widetilde{\theta}_{ij} - \theta_{ij} \in [-\pi, \pi]$ and let $y \in \mathbb{R}$ be a dual variable. The Lagrangian for (\ref{eq:cosine_problem}) is
\begin{align}
    \mathcal{L}(\Theta, y) = \sum_{i \sim j} \cos(\widetilde{\theta}_{ij} - \theta_{ij}) + y \Bigg(\sum_{i\sim j}\theta_{ij} -2k\pi\Bigg)
    \label{eq:cos_lagrangian}
\end{align}
%KKT conditions are
%\begin{align}
%    &\exists \ y\in [-1,1],\; \forall_{i\sim j}  \sin(\widetilde{\theta}_{ij}-\theta_{ij})  = y, % \label{eq:kktcos1}\\
%    &\exists \ k \in \{0,\dots,n-1\}, \; \sum_{i\sim j}\theta_{ij} = 2k\pi.
%    \label{eq:kktcos2}
%\end{align}
%From (\ref{eq:kktcos1}) and (\ref{eq:kktcos2}),
From (\ref{eq:cos_lagrangian}),  we have the sufficient stationarity conditions
\begin{align}
    &\exists \ w\in [-\pi,\pi],\; \forall_{i\sim j} \ \widetilde{\theta}_{ij}-\theta_{ij}  = w,  \label{eq:kktcos3}\\
    &\exists \ k \in \{0,\dots,n-1\}, \; \sum_{i\sim j}\theta_{ij} = 2k\pi.
    \label{eq:kktcos4}
\end{align}
Summing (\ref{eq:kktcos3}) over all the edges of the cycle graph we have
\begin{equation}
    \sum_{i\sim j}\widetilde{\theta}_{ij} - \sum_{i\sim j}\theta_{ij} = n w.
    \label{eq:kktcos5}
\end{equation}
Combining (\ref{eq:kktcos4}) and (\ref{eq:kktcos5}) with $\sum_{i\sim j}\widetilde{\theta}_{ij}=\gamma$ yields
\begin{align}
\theta_{ij} = \widetilde{\theta}_{ij} - \gamma/n + 2k\pi/n.
\label{eq:f_optimal_thetas}
\end{align}
From (\ref{eq:stationary_solution_cycle_so2}), we have $R_i R_j^\top = \widetilde{R}_{ij}E_k^\top$, for $i\sim j$. Thus, 
\begin{equation}
    \angle\big(R_i R_j^\top\big) = \angle\big(\widetilde{R}_{ij}\big) - \angle\big(E_k\big)
\end{equation}
which is simply (\ref{eq:f_optimal_thetas}) since $\angle(E_k) = \gamma/n - 2k\pi / n$.

\section*{Proof of Theorem \ref{theo:spectrum}}
We will prove the result for the spectrum of $\tilde{A}'$, which is equal to that of $\tilde{A}$ since the matrices are similar. We start by showing that the block-vectors $V^k \in \mathbb{R}^{3n\times 3}$, with
\begin{equation}
    V^k := \begin{bmatrix}
    E_k^{0} \\ E_k^{1} \\ \vdots \\ E_k^{{n-1}}
    \end{bmatrix},
    \label{eq:vk}
\end{equation}
indexed by $k\in\{0,\dots,n-1\}$ span invariant subspaces of $\tilde{A}'$ i.e., $\exists\; H \in \mathbb{R}^{3\times 3} : \tilde{A}' V^k = V^k H$. For this step, it suffices to compute $\tilde{A}' V ^k$. From (\ref{eq:Rtildeprime}) and (\ref{eq:vk})
\begin{align}
    \tilde{A}'
    \begin{bmatrix}
    E_k^0 \\
    \vdots \\
    E_k^i \\
    \vdots \\
    E_k^{n-1}
    \end{bmatrix} &= 
    \begin{bmatrix}
    E_k^1 + E^\top E_k^{n-1} \\
    \vdots \\
    E_k^{i-1} + E_k^{i+1} \\
    \vdots \\
    E E_k^0 + E_k^{n-2} \\
    \end{bmatrix} \nonumber \\
    &= \begin{bmatrix}
    E_k^0 \\
    \vdots \\
    E_k^i  \\
    \vdots \\
    E_k^{n-1} \\
    \end{bmatrix}\big(E_k + E_k^\top\big).
    \label{eq:teo5_1}
\end{align}
It compact notation, (\ref{eq:teo5_1}) reads as 
\begin{equation}
    \tilde{A}' V^k = V^k ( E_k + E_k^\top).
    \label{eq:compactinvarsubspace}
\end{equation}
Since $E_k + E_k^\top \in \mathrm{S}^3$, let its spectral decomposition be
\begin{equation}
    E_k + E_k^\top = J \Sigma J^\top, 
    \label{eq:spectruminvarsub}
\end{equation}
with $J\in \mathrm{O}(3)$ and $\Sigma\in\mathbb{R}^{3\times 3}$ diagonal. From (\ref{eq:compactinvarsubspace}) we have
\begin{equation}
    \tilde{A}' (V^k J) = (V^k J) \Sigma.
\end{equation}
The diagonal of $\Sigma$ contains thus three eigenvalues of $\tilde{A}'$. From (\ref{eq:spectruminvarsub}), these eigenvalues are those of $E_k+E_k^\top$ i.e., $\{3, 2\cos(\angle(E_k)), 2\cos(\angle(E_k))\}$.  By definition, $\angle(E_k)=\gamma/n - 2k\pi/n$, thus
\begin{equation}
    \big\{2\cos(\gamma/n - 2k\pi/n)\big\}_{k=0,\dots,n-1} \subset \sigma(\tilde{A}),
    \label{eq:theo5_spec1}
\end{equation}
with each eigenvalue having multiplicity 2.

In order to identify the remaining $n$ eigenvalues of $\widetilde{R}$ let $\hat{n}$ denote the axis of the cycle error $E$ i.e., $E\hat{n}=\hat{n}$ and $E^\top \hat{n} = \hat{n}$. Define the vectors $z^k\in\mathbb{R}^{3n}$
\begin{equation}
    z^k := \begin{bmatrix}
    1 \\
    \cos\big(1\frac{2k\pi}{n}\big) \\
    \cos\big(2\frac{2k\pi}{n}\big) \\
    \vdots \\
    \cos\big((n-1)\frac{2k\pi}{n}\big)
    \end{bmatrix} \otimes \hat{n},
\end{equation}
indexed by $k\in\{0,\dots,n-1\}$. We have
\begin{align}
    &\tilde{A}' z^k = \begin{bmatrix}
    \big(\cos\big(\frac{2k\pi}{n}\big) + \cos\big((n-1)\frac{2k\pi}{n})\big)\hat{n}  \\
    \vdots \\
    \big(\cos\big((i-1)\frac{2k\pi}{n}\big)+ \cos\big((i+1)\frac{2k\pi}{n}\big) \big)\hat{n} \\
    \vdots \\
    \big(\cos\big((n-2)\frac{2k\pi}{n}\big) + \cos\big((n-1)\frac{2k\pi}{n}\big) \big)\hat{n}
    \end{bmatrix} \nonumber \\
    &= \bigg(2\cos\bigg(\frac{2k\pi}{n}\bigg)\bigg)
    \begin{bmatrix}
    1\\
    \vdots \\
    \cos\big(i\frac{2k\pi}{n}\big) \\
    \vdots \\
    \cos\big((n-1)\frac{2k\pi}{n}\big) 
    \end{bmatrix} \otimes \hat{n}
    \label{eq:theo5compact_zk}
\end{align}
In compact notation, (\ref{eq:theo5compact_zk}) reads as
\begin{equation}
    \tilde{A}' z^k = \big(2\cos(2k\pi/n)\big) z^k,
\end{equation}
for $k\in\{0,\dots,n-1\}$. It follows that 
\begin{equation}
    \big\{2\cos(2k\pi/n)\big\}_{k=0,\dots,n-1} \subset \sigma(\tilde{A}).
    \label{eq:theo5_spec2}
\end{equation}
Finally, we have
\begin{align}
    \sigma(\tilde{A}) = &\big\{2\cos\big(\gamma/n - 2k\pi/n\big) \big\}_{k=0,\dots, n-1}\nonumber \\
    &\cup \big\{2\cos\big(2k\pi/n\big)\big\}_{k=0,\dots, n-1}.
\end{align}
from combining (\ref{eq:theo5_spec1}) and (\ref{eq:theo5_spec2}).

% use section* for acknowledgment
\ifCLASSOPTIONcompsoc
  % The Computer Society usually uses the plural form
  \section*{Acknowledgments}
\else
  % regular IEEE prefers the singular form
  \section*{Acknowledgment}
\fi

 This work was supported by  LARSyS funding (DOI: 10.54499/LA/P/0083/2020, 10.54499/UIDP/50009/2020, and 10.54499/UIDB/50009/2020], through Fundação para a Ciência e a Tecnologia. M Marques and J Costeira were also supported by the PT SmartRetail project [PRR - 02/C05-i01.01/2022.PC645440011-00000062], through IAPMEI - Agência para a Competitividade e Inovação.

% Can use something like this to put references on a page
% by themselves when using endfloat and the captionsoff option.
\ifCLASSOPTIONcaptionsoff
  \newpage
\fi

% trigger a \newpage just before the given reference
% number - used to balance the columns on the last page
% adjust value as needed - may need to be readjusted if
% the document is modified later
%\IEEEtriggeratref{8}
% The "triggered" command can be changed if desired:
%\IEEEtriggercmd{\enlargethispage{-5in}}

% references section

% can use a bibliography generated by BibTeX as a .bbl file
% BibTeX documentation can be easily obtained at:
% http://mirror.ctan.org/biblio/bibtex/contrib/doc/
% The IEEEtran BibTeX style support page is at:
% http://www.michaelshell.org/tex/ieeetran/bibtex/
%\bibliographystyle{IEEEtran}
% argument is your BibTeX string definitions and bibliography database(s)
%\bibliography{IEEEabrv,../bib/paper}
%
% <OR> manually copy in the resultant .bbl file
% set second argument of \begin to the number of references
% (used to reserve space for the reference number labels box)
%\begin{thebibliography}{1}
\bibliographystyle{IEEEtran}
\bibliography{bib}

\end{document}

%% file: equivalentcycles.tikz
\tikzset{every picture/.style={line width=0.3pt}} %set default line width to 0.75pt        

\begin{tikzpicture}[x=0.75pt,y=0.75pt,yscale=-1,xscale=1]
%uncomment if require: \path (0,178); %set diagram left start at 0, and has height of 178

%Straight Lines [id:da4091387554883993] 
\draw    (121.66,93.92) -- (125.3,71.83) ;
%Straight Lines [id:da2591325350584336] 
\draw    (121.66,93.92) -- (143.92,75.96) ;
%Straight Lines [id:da7501347623213395] 
\draw    (125.3,71.83) -- (143.92,75.96) ;
%Straight Lines [id:da733052429521652] 
\draw    (121.66,93.92) -- (143.92,96.12) ;
%Straight Lines [id:da08466741149845247] 
\draw    (143.92,75.96) -- (143.92,96.12) ;
%Straight Lines [id:da7570276539389436] 
\draw  [dash pattern={on 0.84pt off 2.51pt}]  (125.3,71.83) -- (125.95,89.55) ;
%Straight Lines [id:da5418310383510282] 
\draw  [dash pattern={on 0.84pt off 2.51pt}]  (121.66,93.92) -- (125.95,89.55) ;
%Straight Lines [id:da14803281682017477] 
\draw  [dash pattern={on 0.84pt off 2.51pt}]  (125.95,89.55) -- (143.92,96.12) ;
%Shape: Parallelogram [id:dp7903763117075648] 
\draw  [fill={rgb, 255:red, 240; green, 240; blue, 240 }  ,fill opacity=1 ] (193.21,10.61) -- (201.13,29.48) -- (185.67,31.84) -- (177.75,12.97) -- cycle ;
%Straight Lines [id:da8809277387236721] 
\draw  [dash pattern={on 0.84pt off 2.51pt}]  (177.75,12.97) -- (206.31,11.34) ;
%Straight Lines [id:da29997151758314744] 
\draw    (193.21,10.61) -- (206.31,11.34) ;
%Straight Lines [id:da24657322697870265] 
\draw    (201.13,29.48) -- (206.31,11.34) ;
%Straight Lines [id:da4122391166004058] 
\draw  [dash pattern={on 0.84pt off 2.51pt}]  (185.67,31.84) -- (206.31,11.34) ;
%Straight Lines [id:da2635377985690075] 
\draw    (177.24,158.56) -- (174.3,135.81) ;
%Straight Lines [id:da9951941777520893] 
\draw    (177.24,158.56) -- (194.65,133.61) ;
%Straight Lines [id:da4420942516722679] 
\draw    (174.3,135.81) -- (194.65,133.61) ;
%Straight Lines [id:da6627654399717321] 
\draw    (177.24,158.56) -- (201.42,153.11) ;
%Straight Lines [id:da8534966331455205] 
\draw    (194.65,133.61) -- (201.42,153.11) ;
%Straight Lines [id:da601378524239594] 
\draw  [dash pattern={on 0.84pt off 2.51pt}]  (174.3,135.81) -- (180.32,152.86) ;
%Straight Lines [id:da8828080119893132] 
\draw  [dash pattern={on 0.84pt off 2.51pt}]  (177.24,158.56) -- (180.32,152.86) ;
%Straight Lines [id:da37329299198704047] 
\draw  [dash pattern={on 0.84pt off 2.51pt}]  (180.32,152.86) -- (201.42,153.11) ;
%Straight Lines [id:da05325232678256531] 
\draw    (264.1,92.76) -- (255.53,71.92) ;
%Straight Lines [id:da07925580568137436] 
\draw    (264.1,92.76) -- (238.2,80.18) ;
%Straight Lines [id:da8038541126348578] 
\draw    (255.53,71.92) -- (238.2,80.18) ;
%Straight Lines [id:da6259461775060541] 
\draw    (264.1,92.76) -- (242.14,100.16) ;
%Straight Lines [id:da8585060079796089] 
\draw    (238.2,80.18) -- (242.14,100.16) ;
%Straight Lines [id:da9764239528032117] 
\draw  [dash pattern={on 0.84pt off 2.51pt}]  (255.53,71.92) -- (258.9,89.44) ;
%Straight Lines [id:da7374971578420049] 
\draw  [dash pattern={on 0.84pt off 2.51pt}]  (264.1,92.76) -- (258.9,89.44) ;
%Straight Lines [id:da5397030917524166] 
\draw  [dash pattern={on 0.84pt off 2.51pt}]  (258.9,89.44) -- (242.14,100.16) ;
%Curve Lines [id:da2267125419527254] 
\draw    (135.44,60.92) .. controls (143.32,46.85) and (154.2,33.4) .. (173.5,30.02) ;
\draw [shift={(133.97,63.61)}, rotate = 298.07] [fill={rgb, 255:red, 0; green, 0; blue, 0 }  ][line width=0.08]  [draw opacity=0] (6.25,-3) -- (0,0) -- (6.25,3) -- cycle    ;
%Curve Lines [id:da9482395087251596] 
\draw    (163.44,145.21) .. controls (147.69,140.45) and (135.32,125.74) .. (132.5,106) ;
\draw [shift={(166.5,146)}, rotate = 192.14] [fill={rgb, 255:red, 0; green, 0; blue, 0 }  ][line width=0.08]  [draw opacity=0] (6.25,-3) -- (0,0) -- (6.25,3) -- cycle    ;
%Curve Lines [id:da5738618865817768] 
\draw    (252.86,114.05) .. controls (249.02,129.87) and (237.18,144.12) .. (216.5,146) ;
\draw [shift={(253.5,111)}, rotate = 100.01] [fill={rgb, 255:red, 0; green, 0; blue, 0 }  ][line width=0.08]  [draw opacity=0] (6.25,-3) -- (0,0) -- (6.25,3) -- cycle    ;
%Curve Lines [id:da07924910640254912] 
\draw    (214.55,29.77) .. controls (238.33,36.3) and (248.13,51.89) .. (250.5,64) ;
\draw [shift={(211.5,29)}, rotate = 12.99] [fill={rgb, 255:red, 0; green, 0; blue, 0 }  ][line width=0.08]  [draw opacity=0] (6.25,-3) -- (0,0) -- (6.25,3) -- cycle    ;
%Straight Lines [id:da10057148938455862] 
\draw [color={rgb, 255:red, 255; green, 0; blue, 0 }  ,draw opacity=1 ][line width=0.75]    (147.95,92) -- (148.08,105.39) ;
\draw [shift={(147.93,90)}, rotate = 89.44] [color={rgb, 255:red, 255; green, 0; blue, 0 }  ,draw opacity=1 ][line width=0.75]    (4.37,-1.32) .. controls (2.78,-0.56) and (1.32,-0.12) .. (0,0) .. controls (1.32,0.12) and (2.78,0.56) .. (4.37,1.32)   ;
%Straight Lines [id:da3348666507056143] 
\draw [color={rgb, 255:red, 0; green, 15; blue, 255 }  ,draw opacity=1 ]   (148.08,105.39) -- (156.78,99.71) ;
\draw [shift={(158.45,98.62)}, rotate = 506.88] [color={rgb, 255:red, 0; green, 15; blue, 255 }  ,draw opacity=1 ][line width=0.75]    (4.37,-1.32) .. controls (2.78,-0.56) and (1.32,-0.12) .. (0,0) .. controls (1.32,0.12) and (2.78,0.56) .. (4.37,1.32)   ;
%Straight Lines [id:da16466644732419755] 
\draw [color={rgb, 255:red, 0; green, 255; blue, 15 }  ,draw opacity=1 ]   (139.13,101.24) -- (148.08,105.39) ;
\draw [shift={(137.32,100.4)}, rotate = 24.83] [color={rgb, 255:red, 0; green, 255; blue, 15 }  ,draw opacity=1 ][line width=0.75]    (4.37,-1.32) .. controls (2.78,-0.56) and (1.32,-0.12) .. (0,0) .. controls (1.32,0.12) and (2.78,0.56) .. (4.37,1.32)   ;
%Straight Lines [id:da8923506697425948] 
\draw [color={rgb, 255:red, 0; green, 15; blue, 255 }  ,draw opacity=1 ]   (250.14,105.52) -- (241.91,101.71) ;
\draw [shift={(240.1,100.87)}, rotate = 384.85] [color={rgb, 255:red, 0; green, 15; blue, 255 }  ,draw opacity=1 ][line width=0.75]    (4.37,-1.32) .. controls (2.78,-0.56) and (1.32,-0.12) .. (0,0) .. controls (1.32,0.12) and (2.78,0.56) .. (4.37,1.32)   ;
%Straight Lines [id:da11172260603003248] 
\draw [color={rgb, 255:red, 255; green, 0; blue, 0 }  ,draw opacity=1 ][line width=0.75]    (248.04,93.67) -- (250.14,105.52) ;
\draw [shift={(247.69,91.7)}, rotate = 79.96] [color={rgb, 255:red, 255; green, 0; blue, 0 }  ,draw opacity=1 ][line width=0.75]    (4.37,-1.32) .. controls (2.78,-0.56) and (1.32,-0.12) .. (0,0) .. controls (1.32,0.12) and (2.78,0.56) .. (4.37,1.32)   ;
%Straight Lines [id:da13982967086673148] 
\draw [color={rgb, 255:red, 0; green, 255; blue, 15 }  ,draw opacity=1 ]   (242.28,111.49) -- (250.14,105.52) ;
\draw [shift={(240.69,112.7)}, rotate = 322.78] [color={rgb, 255:red, 0; green, 255; blue, 15 }  ,draw opacity=1 ][line width=0.75]    (4.37,-1.32) .. controls (2.78,-0.56) and (1.32,-0.12) .. (0,0) .. controls (1.32,0.12) and (2.78,0.56) .. (4.37,1.32)   ;
%Straight Lines [id:da37961362549928024] 
\draw [color={rgb, 255:red, 0; green, 15; blue, 255 }  ,draw opacity=1 ]   (186.14,43.52) -- (181.18,50.02) ;
\draw [shift={(179.97,51.61)}, rotate = 307.31] [color={rgb, 255:red, 0; green, 15; blue, 255 }  ,draw opacity=1 ][line width=0.75]    (4.37,-1.32) .. controls (2.78,-0.56) and (1.32,-0.12) .. (0,0) .. controls (1.32,0.12) and (2.78,0.56) .. (4.37,1.32)   ;
%Straight Lines [id:da042684855078161665] 
\draw [color={rgb, 255:red, 255; green, 0; blue, 0 }  ,draw opacity=1 ][line width=0.75]    (182.74,35.46) -- (186.14,43.52) ;
\draw [shift={(181.97,33.61)}, rotate = 67.19] [color={rgb, 255:red, 255; green, 0; blue, 0 }  ,draw opacity=1 ][line width=0.75]    (4.37,-1.32) .. controls (2.78,-0.56) and (1.32,-0.12) .. (0,0) .. controls (1.32,0.12) and (2.78,0.56) .. (4.37,1.32)   ;
%Straight Lines [id:da7308155467816158] 
\draw [color={rgb, 255:red, 0; green, 255; blue, 15 }  ,draw opacity=1 ]   (186.14,43.52) -- (194.98,42.78) ;
\draw [shift={(196.97,42.61)}, rotate = 535.21] [color={rgb, 255:red, 0; green, 255; blue, 15 }  ,draw opacity=1 ][line width=0.75]    (4.37,-1.32) .. controls (2.78,-0.56) and (1.32,-0.12) .. (0,0) .. controls (1.32,0.12) and (2.78,0.56) .. (4.37,1.32)   ;
%Straight Lines [id:da01078837950351963] 
\draw [color={rgb, 255:red, 0; green, 15; blue, 255 }  ,draw opacity=1 ]   (208.69,160.7) -- (212.96,153.34) ;
\draw [shift={(213.97,151.61)}, rotate = 480.17] [color={rgb, 255:red, 0; green, 15; blue, 255 }  ,draw opacity=1 ][line width=0.75]    (4.37,-1.32) .. controls (2.78,-0.56) and (1.32,-0.12) .. (0,0) .. controls (1.32,0.12) and (2.78,0.56) .. (4.37,1.32)   ;
%Straight Lines [id:da4195461379880244] 
\draw [color={rgb, 255:red, 255; green, 0; blue, 0 }  ,draw opacity=1 ][line width=0.75]    (203.77,149.45) -- (208.69,160.7) ;
\draw [shift={(202.97,147.61)}, rotate = 66.39] [color={rgb, 255:red, 255; green, 0; blue, 0 }  ,draw opacity=1 ][line width=0.75]    (4.37,-1.32) .. controls (2.78,-0.56) and (1.32,-0.12) .. (0,0) .. controls (1.32,0.12) and (2.78,0.56) .. (4.37,1.32)   ;
%Straight Lines [id:da9277044287351993] 
\draw [color={rgb, 255:red, 0; green, 255; blue, 15 }  ,draw opacity=1 ]   (198.97,160.63) -- (208.69,160.7) ;
\draw [shift={(196.97,160.61)}, rotate = 0.41] [color={rgb, 255:red, 0; green, 255; blue, 15 }  ,draw opacity=1 ][line width=0.75]    (4.37,-1.32) .. controls (2.78,-0.56) and (1.32,-0.12) .. (0,0) .. controls (1.32,0.12) and (2.78,0.56) .. (4.37,1.32)   ;
%Straight Lines [id:da6128837034792584] 
\draw    (282,93.89) -- (286.48,71.95) ;
%Straight Lines [id:da2502744193080174] 
\draw    (282,93.89) -- (304.93,76.8) ;
%Straight Lines [id:da2982956938842324] 
\draw    (286.48,71.95) -- (304.93,76.8) ;
%Straight Lines [id:da46868173795279344] 
\draw    (282,93.89) -- (304.8,97.03) ;
%Straight Lines [id:da5377971348863767] 
\draw    (304.93,76.8) -- (304.8,97.03) ;
%Straight Lines [id:da9718923931323425] 
\draw  [dash pattern={on 0.84pt off 2.51pt}]  (286.48,71.95) -- (286.45,89.68) ;
%Straight Lines [id:da5562767348048896] 
\draw  [dash pattern={on 0.84pt off 2.51pt}]  (282,93.89) -- (286.45,89.68) ;
%Straight Lines [id:da4489976486529442] 
\draw  [dash pattern={on 0.84pt off 2.51pt}]  (286.45,89.68) -- (304.8,97.03) ;
%Straight Lines [id:da6175185997703351] 
\draw [color={rgb, 255:red, 255; green, 0; blue, 0 }  ,draw opacity=1 ][line width=0.75]    (366.01,148.61) -- (366.01,161.97) ;
\draw [shift={(366.01,146.61)}, rotate = 90] [color={rgb, 255:red, 255; green, 0; blue, 0 }  ,draw opacity=1 ][line width=0.75]    (4.37,-1.32) .. controls (2.78,-0.56) and (1.32,-0.12) .. (0,0) .. controls (1.32,0.12) and (2.78,0.56) .. (4.37,1.32)   ;
%Straight Lines [id:da1904869841087261] 
\draw [color={rgb, 255:red, 0; green, 15; blue, 255 }  ,draw opacity=1 ]   (366.01,161.97) -- (374.36,155.8) ;
\draw [shift={(375.97,154.61)}, rotate = 503.57] [color={rgb, 255:red, 0; green, 15; blue, 255 }  ,draw opacity=1 ][line width=0.75]    (4.37,-1.32) .. controls (2.78,-0.56) and (1.32,-0.12) .. (0,0) .. controls (1.32,0.12) and (2.78,0.56) .. (4.37,1.32)   ;
%Straight Lines [id:da9843802803940208] 
\draw [color={rgb, 255:red, 0; green, 255; blue, 15 }  ,draw opacity=1 ]   (356.83,158.35) -- (366.01,161.97) ;
\draw [shift={(354.97,157.61)}, rotate = 21.52] [color={rgb, 255:red, 0; green, 255; blue, 15 }  ,draw opacity=1 ][line width=0.75]    (4.37,-1.32) .. controls (2.78,-0.56) and (1.32,-0.12) .. (0,0) .. controls (1.32,0.12) and (2.78,0.56) .. (4.37,1.32)   ;
%Straight Lines [id:da5779238624960554] 
\draw    (336.07,155.28) -- (340.04,132.61) ;
%Straight Lines [id:da47220643650412475] 
\draw    (336.07,155.28) -- (359.13,137.06) ;
%Straight Lines [id:da10368573943355874] 
\draw    (340.04,132.61) -- (359.13,137.06) ;
%Straight Lines [id:da4401740985404897] 
\draw    (336.07,155.28) -- (359.58,157.84) ;
%Straight Lines [id:da9170664480353997] 
\draw    (359.13,137.06) -- (359.58,157.84) ;
%Straight Lines [id:da4693474602855] 
\draw  [dash pattern={on 0.84pt off 2.51pt}]  (340.04,132.61) -- (340.52,150.83) ;
%Straight Lines [id:da6012220381997443] 
\draw  [dash pattern={on 0.84pt off 2.51pt}]  (336.07,155.28) -- (340.52,150.83) ;
%Straight Lines [id:da06611846136661725] 
\draw  [dash pattern={on 0.84pt off 2.51pt}]  (340.52,150.83) -- (359.58,157.84) ;
%Straight Lines [id:da4621641039096178] 
\draw    (403.07,96.28) -- (407.04,73.61) ;
%Straight Lines [id:da12210897974388613] 
\draw    (403.07,96.28) -- (426.13,78.06) ;
%Straight Lines [id:da4171356517218121] 
\draw    (407.04,73.61) -- (426.13,78.06) ;
%Straight Lines [id:da16121717370681277] 
\draw    (403.07,96.28) -- (426.58,98.84) ;
%Straight Lines [id:da08882484148491554] 
\draw    (426.13,78.06) -- (426.58,98.84) ;
%Straight Lines [id:da22253660569679745] 
\draw  [dash pattern={on 0.84pt off 2.51pt}]  (407.04,73.61) -- (407.52,91.83) ;
%Straight Lines [id:da0060322288764824705] 
\draw  [dash pattern={on 0.84pt off 2.51pt}]  (403.07,96.28) -- (407.52,91.83) ;
%Straight Lines [id:da43752095393400414] 
\draw  [dash pattern={on 0.84pt off 2.51pt}]  (407.52,91.83) -- (426.58,98.84) ;
%Straight Lines [id:da4297805309352738] 
\draw [color={rgb, 255:red, 255; green, 0; blue, 0 }  ,draw opacity=1 ][line width=0.75]    (430.01,92.61) -- (430.01,105.97) ;
\draw [shift={(430.01,90.61)}, rotate = 90] [color={rgb, 255:red, 255; green, 0; blue, 0 }  ,draw opacity=1 ][line width=0.75]    (4.37,-1.32) .. controls (2.78,-0.56) and (1.32,-0.12) .. (0,0) .. controls (1.32,0.12) and (2.78,0.56) .. (4.37,1.32)   ;
%Straight Lines [id:da4271794116426777] 
\draw [color={rgb, 255:red, 0; green, 15; blue, 255 }  ,draw opacity=1 ]   (430.01,105.97) -- (438.36,99.8) ;
\draw [shift={(439.97,98.61)}, rotate = 503.57] [color={rgb, 255:red, 0; green, 15; blue, 255 }  ,draw opacity=1 ][line width=0.75]    (4.37,-1.32) .. controls (2.78,-0.56) and (1.32,-0.12) .. (0,0) .. controls (1.32,0.12) and (2.78,0.56) .. (4.37,1.32)   ;
%Straight Lines [id:da1324663185320224] 
\draw [color={rgb, 255:red, 0; green, 255; blue, 15 }  ,draw opacity=1 ]   (420.83,102.35) -- (430.01,105.97) ;
\draw [shift={(418.97,101.61)}, rotate = 21.52] [color={rgb, 255:red, 0; green, 255; blue, 15 }  ,draw opacity=1 ][line width=0.75]    (4.37,-1.32) .. controls (2.78,-0.56) and (1.32,-0.12) .. (0,0) .. controls (1.32,0.12) and (2.78,0.56) .. (4.37,1.32)   ;
%Curve Lines [id:da09042146601562073] 
\draw    (294.66,61.24) .. controls (300.08,45.66) and (311.2,33.4) .. (330.5,30.02) ;
\draw [shift={(293.69,64.26)}, rotate = 286.39] [fill={rgb, 255:red, 0; green, 0; blue, 0 }  ][line width=0.08]  [draw opacity=0] (6.25,-3) -- (0,0) -- (6.25,3) -- cycle    ;
%Curve Lines [id:da9379739644847175] 
\draw    (378.56,30.77) .. controls (402.49,37.38) and (414.13,53.89) .. (416.5,66) ;
\draw [shift={(375.5,30)}, rotate = 12.99] [fill={rgb, 255:red, 0; green, 0; blue, 0 }  ][line width=0.08]  [draw opacity=0] (6.25,-3) -- (0,0) -- (6.25,3) -- cycle    ;
%Curve Lines [id:da5770313158445007] 
\draw    (414.86,114.98) .. controls (411.02,130.81) and (399.18,145.06) .. (378.5,146.94) ;
\draw [shift={(415.5,111.94)}, rotate = 100.01] [fill={rgb, 255:red, 0; green, 0; blue, 0 }  ][line width=0.08]  [draw opacity=0] (6.25,-3) -- (0,0) -- (6.25,3) -- cycle    ;
%Straight Lines [id:da1792335168314768] 
\draw    (338.07,34.28) -- (342.04,11.61) ;
%Straight Lines [id:da09981342824180783] 
\draw    (338.07,34.28) -- (361.13,16.06) ;
%Straight Lines [id:da14051917780559164] 
\draw    (342.04,11.61) -- (361.13,16.06) ;
%Straight Lines [id:da45696780537452764] 
\draw    (338.07,34.28) -- (361.58,36.84) ;
%Straight Lines [id:da6795296612416092] 
\draw    (361.13,16.06) -- (361.58,36.84) ;
%Straight Lines [id:da8584998388738709] 
\draw  [dash pattern={on 0.84pt off 2.51pt}]  (342.04,11.61) -- (342.52,29.83) ;
%Straight Lines [id:da13348837344990538] 
\draw  [dash pattern={on 0.84pt off 2.51pt}]  (338.07,34.28) -- (342.52,29.83) ;
%Straight Lines [id:da39217759327590096] 
\draw  [dash pattern={on 0.84pt off 2.51pt}]  (342.52,29.83) -- (361.58,36.84) ;
%Straight Lines [id:da3571789973510948] 
\draw [color={rgb, 255:red, 255; green, 0; blue, 0 }  ,draw opacity=1 ][line width=0.75]    (365.01,30.61) -- (365.01,43.97) ;
\draw [shift={(365.01,28.61)}, rotate = 90] [color={rgb, 255:red, 255; green, 0; blue, 0 }  ,draw opacity=1 ][line width=0.75]    (4.37,-1.32) .. controls (2.78,-0.56) and (1.32,-0.12) .. (0,0) .. controls (1.32,0.12) and (2.78,0.56) .. (4.37,1.32)   ;
%Straight Lines [id:da30900877127789783] 
\draw [color={rgb, 255:red, 0; green, 15; blue, 255 }  ,draw opacity=1 ]   (365.01,43.97) -- (373.36,37.8) ;
\draw [shift={(374.97,36.61)}, rotate = 503.57] [color={rgb, 255:red, 0; green, 15; blue, 255 }  ,draw opacity=1 ][line width=0.75]    (4.37,-1.32) .. controls (2.78,-0.56) and (1.32,-0.12) .. (0,0) .. controls (1.32,0.12) and (2.78,0.56) .. (4.37,1.32)   ;
%Straight Lines [id:da8660565533107607] 
\draw [color={rgb, 255:red, 0; green, 255; blue, 15 }  ,draw opacity=1 ]   (355.83,40.35) -- (365.01,43.97) ;
\draw [shift={(353.97,39.61)}, rotate = 21.52] [color={rgb, 255:red, 0; green, 255; blue, 15 }  ,draw opacity=1 ][line width=0.75]    (4.37,-1.32) .. controls (2.78,-0.56) and (1.32,-0.12) .. (0,0) .. controls (1.32,0.12) and (2.78,0.56) .. (4.37,1.32)   ;
%Straight Lines [id:da5004056171628081] 
\draw [color={rgb, 255:red, 208; green, 2; blue, 27 }  ,draw opacity=1 ]   (282,93.89) -- (308.69,90.26) ;
%Straight Lines [id:da03332859462285287] 
\draw [color={rgb, 255:red, 208; green, 2; blue, 27 }  ,draw opacity=1 ]   (282,93.89) -- (297.17,98) ;
%Straight Lines [id:da9126110562873514] 
\draw [color={rgb, 255:red, 208; green, 2; blue, 27 }  ,draw opacity=1 ]   (308.69,90.26) -- (308.69,106.26) ;
%Straight Lines [id:da5369300307589755] 
\draw [color={rgb, 255:red, 208; green, 2; blue, 27 }  ,draw opacity=1 ]   (282,93.89) -- (297.17,116) ;
%Straight Lines [id:da2574209067946822] 
\draw [color={rgb, 255:red, 208; green, 2; blue, 27 }  ,draw opacity=1 ]   (308.69,106.26) -- (297.17,116) ;
%Straight Lines [id:da6740110304836924] 
\draw [color={rgb, 255:red, 208; green, 2; blue, 27 }  ,draw opacity=1 ]   (297.17,98) -- (297.17,116) ;
%Straight Lines [id:da14851687319217366] 
\draw [color={rgb, 255:red, 208; green, 2; blue, 27 }  ,draw opacity=1 ]   (308.69,90.26) -- (297.17,98) ;
%Straight Lines [id:da013785681205173539] 
\draw [color={rgb, 255:red, 208; green, 2; blue, 27 }  ,draw opacity=1 ] [dash pattern={on 0.84pt off 2.51pt}]  (308.69,106.26) -- (282,93.89) ;
%Curve Lines [id:da8454851518115641] 
\draw    (325.63,146.56) .. controls (308.63,142.35) and (300.38,134.63) .. (296.6,124.24) ;
\draw [shift={(328.69,147.26)}, rotate = 191.91] [fill={rgb, 255:red, 0; green, 0; blue, 0 }  ][line width=0.08]  [draw opacity=0] (6.25,-3) -- (0,0) -- (6.25,3) -- cycle    ;

% Text Node
\draw (164,84) node [anchor=north west][inner sep=0.75pt]  [font=\footnotesize]  {$R_{1}$};
% Text Node
\draw (187,107) node [anchor=north west][inner sep=0.75pt]  [font=\footnotesize]  {$R_{2}$};
% Text Node
\draw (209.3,84) node [anchor=north west][inner sep=0.75pt]  [font=\footnotesize]  {$R_{3}$};
% Text Node
\draw (187,60) node [anchor=north west][inner sep=0.75pt]  [font=\footnotesize]  {$R_{4}$};
% Text Node
\draw (128,142) node [anchor=north west][inner sep=0.75pt]  [font=\footnotesize]  {$\widetilde{R}_{12}$};
% Text Node
\draw (128,20) node [anchor=north west][inner sep=0.75pt]  [font=\footnotesize]  {$\widetilde{R}_{41}$};
% Text Node
\draw (240,20) node [anchor=north west][inner sep=0.75pt]  [font=\footnotesize]  {$\widetilde{R}_{34}$};
% Text Node
\draw (290,145) node [anchor=north west][inner sep=0.75pt]  [font=\footnotesize]  {$I$};
% Text Node
\draw (347,107) node [anchor=north west][inner sep=0.75pt]  [font=\footnotesize]  {$R'_{2}$};
% Text Node
\draw (372.01,80.97) node [anchor=north west][inner sep=0.75pt]  [font=\footnotesize]  {$R'_{3}$};
% Text Node
\draw (346.95,60) node [anchor=north west][inner sep=0.75pt]  [font=\footnotesize]  {$R'_{4}$};
% Text Node
\draw (406.55,145) node [anchor=north west][inner sep=0.75pt]  [font=\footnotesize]  {$I$};
% Text Node
\draw (407.55,22) node [anchor=north west][inner sep=0.75pt]  [font=\footnotesize]  {$I$};
% Text Node
\draw (282,22) node [anchor=north west][inner sep=0.75pt]  [font=\footnotesize,color={rgb, 255:red, 208; green, 2; blue, 27 }  ,opacity=1 ]  {$R_{z}( \gamma )$};
% Text Node
\draw (322,82) node [anchor=north west][inner sep=0.75pt]  [font=\footnotesize]  {$R'_{1}$};
% Text Node
\draw (240,142) node [anchor=north west][inner sep=0.75pt]  [font=\footnotesize]  {$\widetilde{R}_{23}$};

\end{tikzpicture}